\documentclass[lettersize,journal]{IEEEtran} 
\usepackage[noadjust]{cite}
\usepackage{amsmath, amsfonts, amssymb}
\usepackage{algorithmic}
\usepackage{graphicx}
\usepackage{textcomp}
\usepackage{xcolor}
\usepackage{caption}
\usepackage{subcaption}
\usepackage{algorithm}
\usepackage{array}
\usepackage{stfloats}
\usepackage{url}
\usepackage{verbatim}
\usepackage{pifont}
\usepackage[normalem]{ulem}
\usepackage{makecell}
\usepackage{array}

\usepackage[table,xcdraw]{xcolor}

\definecolor{dark_green}{rgb}{0, 0.5, 0}
\definecolor{dark_yellow}{rgb}{0.75, 0.75, 0}

\usepackage{etoolbox}
\makeatletter 
\pretocmd\@bibitem{\color{black}\csname keycolor#1\endcsname}{}{\fail}
\newcommand\citecolor[1]{\@namedef{keycolor#1}{\color{blue}}} 
\makeatother

\begin{document}

\setlength{\abovedisplayskip}{5pt}
\setlength{\belowdisplayskip}{5pt}

\title{NeuromorphicRx: From Neural to Spiking Receiver}

\author{Ankit Gupta,~\IEEEmembership{Member,~IEEE,} Onur Dizdar,~\IEEEmembership{Senior Member,~IEEE,} Yun Chen,~\IEEEmembership{Member,~IEEE}, Fehmi Emre Kadan,~\IEEEmembership{Member,~IEEE,} Ata Sattarzadeh,~\IEEEmembership{Member,~IEEE,} and Stephen Wang,~\IEEEmembership{Senior Member,~IEEE.}
\thanks{The authors are with VIAVI Marconi Labs, VIAVI Solutions Inc., Stevenage SG1 2AN, UK. (e-mail: \{ankit.gupta, onur.dizdar, yun.chen, fehmiemre.kadan, ata.sattarzadeh, stephen.wang\}@viavisolutions.com.)}}


\IEEEpubid{0000--0000/00\$00.00~\copyright~2021 IEEE}

\maketitle

\begin{abstract}
In this work, we propose a novel energy-efficient spiking neural network (SNN)-based receiver for 5G-NR OFDM system, called neuromorphic receiver (NeuromorphicRx), replacing the channel estimation, equalization and symbol demapping blocks. We leverage domain knowledge to design the input with spiking encoding and propose a deep convolutional SNN with spike-element-wise residual connections. We integrate an SNN with artificial neural network (ANN) hybrid architecture to obtain soft outputs and employ surrogate gradient descent for training. We focus on generalization across diverse scenarios and robustness through quantized aware training. We focus on interpretability of NeuromorphicRx for 5G-NR signals and perform detailed ablation study for 5G-NR signals. Our extensive numerical simulations show that NeuromorphicRx is capable of achieving significant block error rate performance gain compared to 5G-NR receivers and similar performance compared to its ANN-based counterparts with $7.6\times$ less energy consumption.
\end{abstract}

\vspace{-0.5cm}
\section{Introduction}

\IEEEPARstart{N}{eural} receivers (NeuralRx) replace multiple signal processing blocks, including channel estimation and interpolation, equalization, and symbol demapping by a single neural network (NN)~\cite{3GPP38843, Hoydis2021, Dorner2018, Honkala2021, Cammerer2020, Gupta2023, Faycal2022, Pihlajasalo2023, Raviv2023, Xie2024, Mei2024, Sun2024, Korpi2023, Wan23, Gupta2022, Gupta2023_2, Gupta2021_2, Xiao2025, Li2025}. The NeuralRx deals with the channel effects and hardware impairments with higher accuracy as the joint training of the multiple blocks enables the NN to remove the inherent design assumptions made in the signal-processing blocks for mathematical tractability, leading to significant performance gains compared to receivers with separate signal processing blocks.

Artificial NNs (ANNs) are widely employed to achieve NN-based processing for wireless communications. Recently, Neuromorphic or Spiking Neural Networks (SNNs) have appeared as a low-power substitute for the ANN-based frameworks. SNNs replace the conventional neurons in ANNs with spiking neurons. Consequently, SNNs employ discrete temporal $1$-bit spikes to encode the data whereas ANNs employ real-value numbers. 
In this sense, the SNNs resemble the human brain more than ANNs and are realized via neuromorphic computing \cite{Hyeryung2021, Skatchkovsky2021_2, Rajendran2019}. Thereby, SNNs can reduce the energy consumption up to $10-20$ times compared to traditional ANN architectures \cite{Ge2023, Ortiz2024, Liu2024, Liu2024_2, Chen2023, Chen2024, Borsos2022, Velusamy2023, Vogginger2022, Wen2024, Chen2023_2, Dakic2024, Xie2022, Hamedani2020, Hamedani2021}. Each spike in an SNN consumes only several picojoules of energy, practically proportional to the volume of spikes undergoing processing \cite{Rajendran2019}. Owing to their benefits, SNNs are employed in wireless communications, such as Integrated Sensing and Communications (ISAC) \cite{Skatchkovsky2021_2, Vogginger2022, Wen2024, Chen2023_2}, semantic communications with ISAC \cite{Chen2023}, sum-rate maximization \cite{Ge2023}, spectrum sensing \cite{Hamedani2021, Liu2024, Hamedani2020, Dakic2024_2}, satellite communications \cite{Ortiz2024, Dakic2024}, distributed wireless networks \cite{Borsos2022}, and distributed routing \cite{Velusamy2023}. 
However, the benefit of energy-efficiency in SNNs comes with the cost of performance degradation compared to ANNs. One reason for this is that the research on SNNs is still in its infancy, with many open problems, such as the best way to train the non-differentiable spikes, how to create a deep SNN, choosing activation functions, spike encoding, and spike reset mechanisms.
\vspace{-0.34cm}
\subsection{Related Work}
Designing a NeuralRx by employing an ANN is a widely investigated topic in literature \cite{Honkala2021, Cammerer2020, Gupta2023, Faycal2022, Pihlajasalo2023, Raviv2023, Xie2024, Mei2024, Sun2024, Korpi2023, Wan23, Gupta2022, Gupta2023_2, Gupta2021_2, Xiao2025, Li2025}. One can broadly classify the related works based on their training methodology, {\sl i.e.}, receiver-side and end-to-end (E2E). The works on receiver-side employ a NeuralRx for the transmitted signal that contains pilots. E2E training enables pilotless OFDM transmissions, achieving additional throughput by utilizing a NeuralRx with either (1) an NN replacing signal mapping block to generate a custom constellation at the transmitter, or (2) superimposed pilots (SIP)~\cite{Faycal2022, Cammerer2020}. The drawback of the SIP approach is the intra-/inter-layer interference in MIMO-OFDM systems, which is addressed in recent studies~\cite{Xiao2025, Li2025} by expert-knowledge interference-cancellation-based NeuralRx. Specifically,~\cite{Xiao2025} leverages symbol-aided channel estimation, fixed-power SIP, and scalable layer/MCS design, while~\cite{Li2025} replaces the LMMSE-based channel estimation with either a variational message passing or ANN. However, the drawback of all abovementioned applications of ANN-based NeuralRx is their high power consumption, which limits their practical deployment in 6G devices with strict power consumption requirements, such as user equipment and battery-powered Internet-of-Things (IoT) devices.
Indeed, 3GPP also raises this concern in~\cite{3GPP38843}, and thus focuses on use cases which require smaller AI modules by replacing only a single signal processing block.  This motivates the use of SNNs for NeuralRx that can provide more ``intelligence-per-joule".

There is a limited number of works on the use of SNNs for designing the NeuralRxs \cite{Liu2024_2, Chen2023_2}. 
In \cite{Liu2024_2}, an SNN-based symbol detector for MIMO-OFDM systems is proposed, where the symbol detection problem is treated as a regression problem. An SNN module is employed at the receiver for channel estimation, equalization, and symbol de-mapping. The SNN is trained using a knowledge distillation-based teacher-student learning algorithm, such that an ANN-based Echo State Network is used as the teacher and SNN-based Liquid State Machine (reservoir computing) is the student. Although reservoir computing training is highly energy-efficient because training is performed only for the output layer, its learning capability and interpretability remain limited due to the fixed and random nature of the reservoir, respectively. The authors in \cite{Chen2023_2} propose neuromorphic ISAC, where an SNN is deployed at the receiver to decode digital data and detect the radar target. The transmission is performed in the form of neuromorphic communications by using impulse radio (IR) transmission and pulse position modulation (PPM) method for symbol mapping. An SNN is employed to provide two binary values as output: (i) for information bits {0, 1}, and (ii) for target detection (presence/absence). However, the authors do not consider conventional waveforms ({\sl e.g.,} OFDM) or modulation schemes ({\sl e.g.,} Quadrature Amplitude Modulation (QAM)) that are widely used in practical systems for transmission.

Although the abovementioned works investigate several aspects of the use of SNNs for neural receivers, there is still a lack of a comprehensive study of SNNs for neural receivers in terms of design, architectures, training methods, generalizability aspects, and performance improvements achieved for practical systems such as 5G New Radio (5G-NR). 
\vspace{-0.15cm}
\subsection{Contributions}
In this work, we propose a novel neural receiver architecture, called NeuromorphicRx, to replace the channel estimation, interpolation, equalization and symbol demapping blocks jointly at multi-antenna receivers. We formulate the symbol detection problem for 5G-NR signals as a multi-label classification problem and output log-likelihood ratios (LLRs) compatible with standard soft channel decoders. 
To the best of the authors' knowledge, this is the first neural receiver architecture that employs SNNs for signal detection in 5G-NR OFDM-based systems with multiple antennas. 

The contributions of the paper are listed as follows:
\begin{enumerate}
    \item We propose an architecture that leverages the domain knowledge to design a real-valued input encoding layer for QAM signals to minimize time-steps and energy consumption, and overcome the limitations of traditional spike-based encoding methods. 
    Furthermore, we input 5G-NR resource grid by passing the whole grid in a slot and associated DMRS as input, which allows learning channel estimation from DMRS and data symbols during the training to improve the performance. 
    \item We propose a novel deep spiking residual networks (ResNet) structure with hybrid readout layer that combines SNN and ANN by employing the SNN in all layers except the last layer, which is designed using an ANN with a sigmoid activation. Moreover, spike-element-wise (SEW) ResNet blocks are used instead of traditional ResNet blocks to achieve identity mapping and tackle the vanishing gradient problem. We show that the proposed ResNet block enhances symbol detection performance.
    \item The proposed NeuromorphicRx is trained by surrogate gradient descent (SGD) and designed to be generalizable and robust for deployment in various environments. Specifically, we focus on the domain-aware generalizability during training in terms of signal-to-noise ratio (SNR), Doppler, delay, TDL/CDL channel models, and various DMRS configurations such that the network can adapt to various environments and settings without any additional training. Furthermore,  we focus on the robustness of NeuromorphicRx for deployment on small mobile devices. Accordingly, we propose ``quantize-aware'' training, where the weights are quantized from full-precision during the training itself to achieve robust performance under quantization. 
    \item We perform an extensive domain-aware ablation study to have an in-depth understanding of the performance of NeuromorphicRx. First, we investigate the activation probabilities and membrane potential for varying SNR, Doppler, and TDL/CDL channel profiles to understand the neuron activation dynamics in wireless networks. Then, we determine the best activation neurons (Lapicque $<$ Leaky $<$ RLeaky), time-steps ($T=2$), surrogate functions (SSO $\approx$ LSO $\approx$ SFS $\approx$ Sigmoid $<$ Fast Sigmoid $<$ ArcTan), SEW ResNet block and its combining operations (AND $<$ IAND $<$ ADD) to decode 5G-NR signals.
    \item We show for the first time that SNN-based receivers can achieve performance as good as their ANN-based counterparts with around $7.6\times$ less energy consumption. Furthermore, we demonstrate that the proposed NeuromorphicRx and training method achieves a significant communications performance gain compared to 5G-NR-based LS/LMMSE receivers, which motivates its deployment in practical systems. 
\end{enumerate}

The organization of the paper is as follows. Section II describes the system model. We describe the proposed architecture and the training method in Section III. Section IV presents the ablation study and numerical results are performed in Section V. Section VI concludes the paper.

\textit{Notation:}
Matrices and vectors are denoted by bold uppercase and lowercase letters, respectively. The operations $|.|$ and $||.||$ denote the absolute
value of a scalar and l2-norm of a vector, respectively. 
Logarithms are natural logarithms, $\log(.) = \log_{e} (.)$  
and $\lfloor\cdot\rceil$ is round-to-nearest integer operation. $\mathbb{C}$ and $\mathbb{R}$ denote the complex and real numbers, respectively.
\vspace{-0.1cm}
\section{System Model}
\label{sec:system_model}
\begin{figure}[t!]
    \centering
    \includegraphics[scale=0.5]{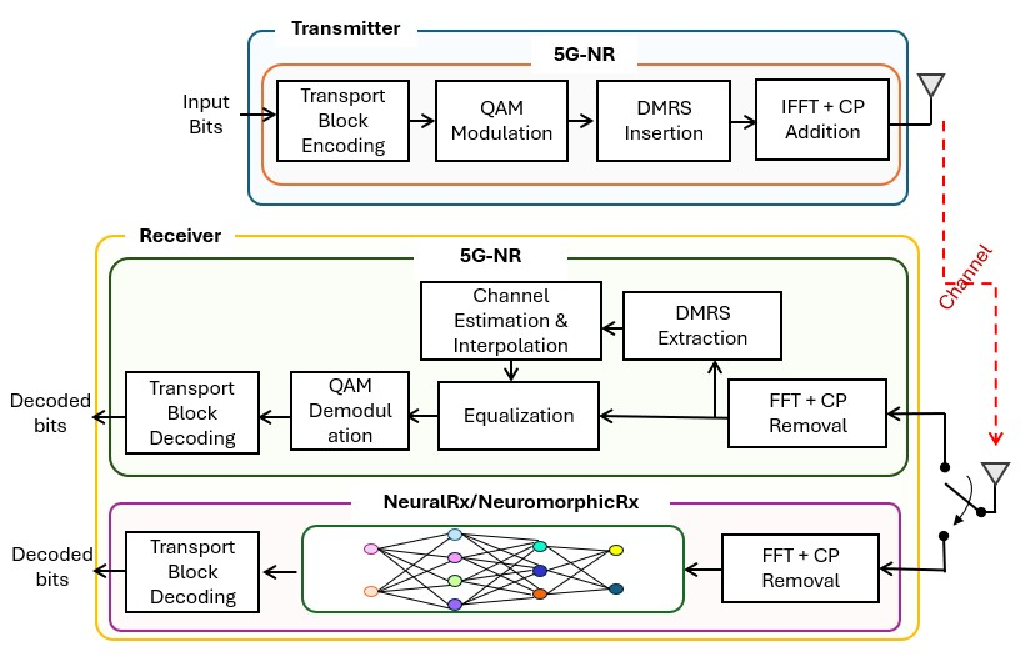}
    \caption{Block diagram of 5G-NR transceiver and NeuralRx/NeuromorphicRx.}
    \vspace{-0.5cm}
   \label{fig:architecture_conv}
\end{figure}

As shown in Fig.~\ref{fig:architecture_conv}, we consider a single-user 5G-NR physical layer uplink shared channel (PUSCH) or single-layer physical layer downlink shared channel (PDSCH) scenario, where the receiver has $N_{R}$ receive antennas.

\subsection{Signal Transmission-Reception}
At the transmitter, the information bits are fed into the transport block encoder to output a series of codewords. Next, complex baseband symbols are created using the symbol mapping block, which is mapped to the Physical Resource Blocks (PRBs) in a Transmission Time Interval (TTI), named as the resource grid, as shown in Fig.~\ref{fig:BW_SNN}a. Additionally, pilots, also referred to as Demodulation Reference Signal (DMRS) are injected into specifically defined subcarriers and OFDM symbols. Next, the PRBs are input to an inverse fast Fourier transform (IFFT) block, which turns complex baseband symbols into time-domain OFDM symbols. Finally, a cyclic prefix (CP) is appended at the beginning of every OFDM symbol to reduce inter-symbol interference.

The obtained signal propagates through the 3GPP TR38.901 TDL/CDL channels and is distorted by Additive White Gaussian Noise (AWGN) at the receiver. The receiver removes the CP and applies a fast Fourier transform (FFT) on each OFDM symbol. The received signal can be expressed as
\begin{align}
\mathbf{y}_{m,n} = \mathbf{h}_{m,n} x_{m,n} + \mathbf{z}_{m,n},
\label{eqn:received}
\end{align}
where $x_{m,n}\in\mathbb{C}$ and $\mathbf{y}_{m,n}\in\mathbb{C}^{N_R\times 1}$ denote the transmitted and received signals, respectively, $\mathbf{h}_{m,n}\in\mathbb{C}^{N_R\times 1}$ denotes the effective (precoded) channel between the BS and UE, and $z_{m,n} \sim \mathcal{CN}(0,N_0) \in\mathbb{C}^{N_R\times 1}$ is the AWGN at the $m$-th OFDM symbol and $n$-th subcarrier for $m\in\left\{0,\ \ldots,\ \ M-1\right\}$ and $n\in\left\{0,\ldots,\ \ N-1\right\}$.
\vspace{-0.1cm}
\subsection{5G-NR Receiver Processing}
At the receiver, the channel estimation is performed after FFT operation to determine the channel estimate by employing the known pilot/DMRS symbols, $p_{i,j}\in\mathbb{C}$, where $i\in \mathcal{I} \subseteq \left\{0,\ \ldots,\ \ M-1\right\}$ and $j\in\mathcal{J} \subseteq\left\{0,\ldots,\ \ N-1\right\}$ represent the OFDM symbol and subcarrier that the pilot is located in, respectively, and  $\mathcal{I}$ and $\mathcal{J}$ are the sets that contain the symbol and subcarrier indexes of DMRS symbols, respectively. We consider the LS channel estimate at DMRS symbols, given as
\begin{align}
\mathbf{\hat{h}}_{i,j}\hspace{-0.05cm}=\hspace{-0.05cm}\mathbf{y}_{i,j} \dfrac{p_{i,j}^\ast}{\left\vert p_{i,j}\right\vert^2}\hspace{-0.05cm} =\hspace{-0.05cm} \mathbf{h}_{i,j} \hspace{-0.05cm}+ \hspace{-0.05cm}\mathbf{\widetilde{h}}_{i,j}, \
\sigma_{i,j}^2 \hspace{-0.05cm}= \hspace{-0.05cm}\mathbb{E}\left[\mathbf{\widetilde{h}^{H}}_{i,j}\mathbf{\widetilde{h}}_{i,j}\right], 
\end{align}
where $\mathbf{\hat{h}}_{i,j}\in\ \mathbb{C}^{N_R\times 1}$ and $\mathbf{\widetilde{h}}_{i,j}\in\ \mathbb{C}^{N_R\times 1}$ denote the estimated effective channel and the estimation error at the DMRS symbols, respectively, and $\sigma_{i,j}^2\in\mathbb{R}$ denotes the estimation error variance. Next, interpolation is applied to obtain the channel estimates and error variances throughout the whole resource grid. In this work, we consider LS channel estimation with a low-complexity linear interpolation (LS channel estimation) and a high-complexity Linear Minimum Mean Square Error (LMMSE) interpolation (LMMSE channel estimation).

Next, we perform LMMSE equalization on each data symbol $\mathbf{y}_{m^\prime n}$, $m^\prime \in \left\{0,\ \ldots,\ \ M-1\right\}\setminus\mathcal{I}$, to determine the estimated data symbols as
\begin{align}
\hat{x}_{m^\prime ,n} = \left(\mathbf{\hat{h}}_{m^\prime ,n}^{H} \mathbf{\hat{h}}_{m^\prime ,n} + \ \sigma_{m^\prime, n}^2\right)^{-1}\mathbf{\hat{h}}_{m^\prime ,n}^{H}\mathbf{y}_{m^\prime ,n}.
\end{align}
Finally, a signal de-mapper calculates the LLRs from the equalized symbols $\hat{x}_{m^\prime ,n}$. Let us denote the $l$-th bit of the symbol at $m^\prime$-th OFDM symbol and $n$-th subcarrier by $b^{l}_{m^{\prime},n}$, $l\in\{0,\ \ldots,\ B_t-1\}$. One can obtain the LLR for $b^{l}_{m^{\prime},n}$ as 
\begin{align}
LLR_{m^\prime ,n}^l \!=\! \log{\!\left(\!{\Pr{\left(\!b_{m^\prime ,n}^l\!=\!1\!\right|{\hat{x}}_{m^\prime ,n}^l\!)}}\big/{\Pr{\left(\!b_{m^\prime ,n}^l\!=\!0\right|{\hat{x}}_{m^\prime ,n}^l\!)\!}}\right)} \nonumber
\end{align}

\begin{figure}[t!]
    \centering
    \begin{subfigure}{0.18\textwidth}
        \centering
        \includegraphics[width=0.9\linewidth]{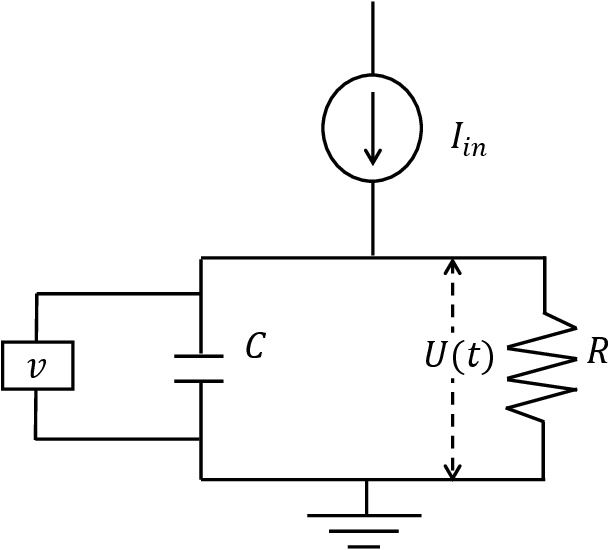}
        \vspace{1cm}
        \caption{RC circuit representation of the LIF neuron.}
        \label{fig:circuit}
    \end{subfigure}
    \begin{subfigure}{0.3\textwidth}
        \centering
        \includegraphics[width=0.97\linewidth]{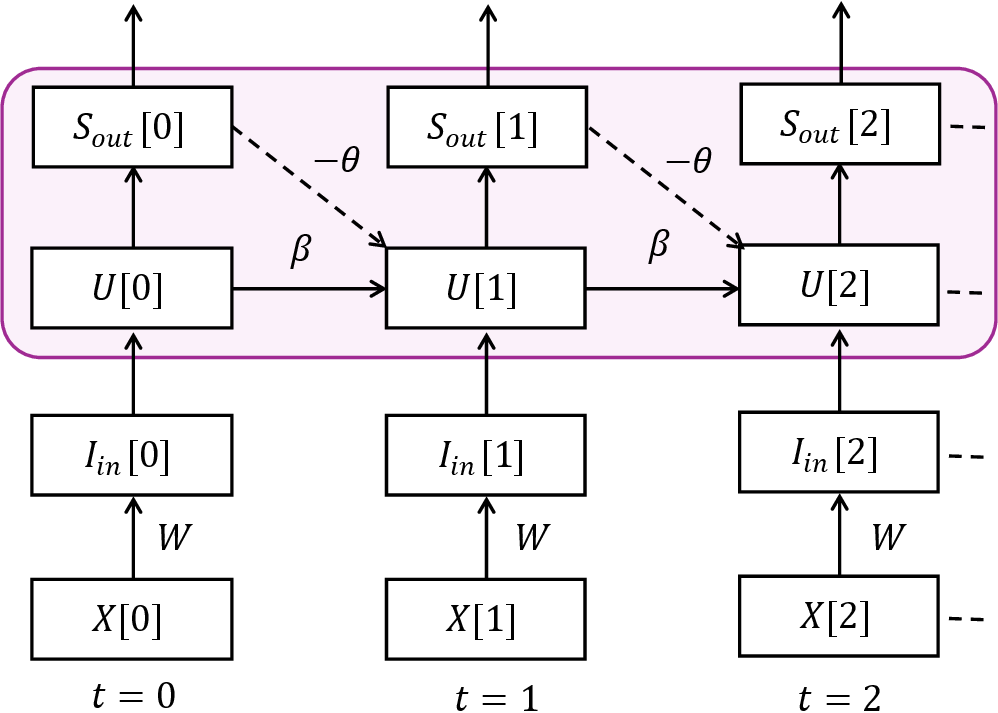}
        \caption{Computational graph of the LIF unrolled over time.}
        \label{fig:graph}
    \end{subfigure}
    \caption{Representation of the LIF neuron.}
    \vspace{-0.35cm}
\end{figure}
\section{Proposed SNN-based Neuromorphic Receiver}
In this section, we first provide fundamentals on spiking neurons, and then describe the SNN-based NeuromorphicRx.

\begin{figure*}[t!]
    \centering
    \includegraphics[scale=0.6]{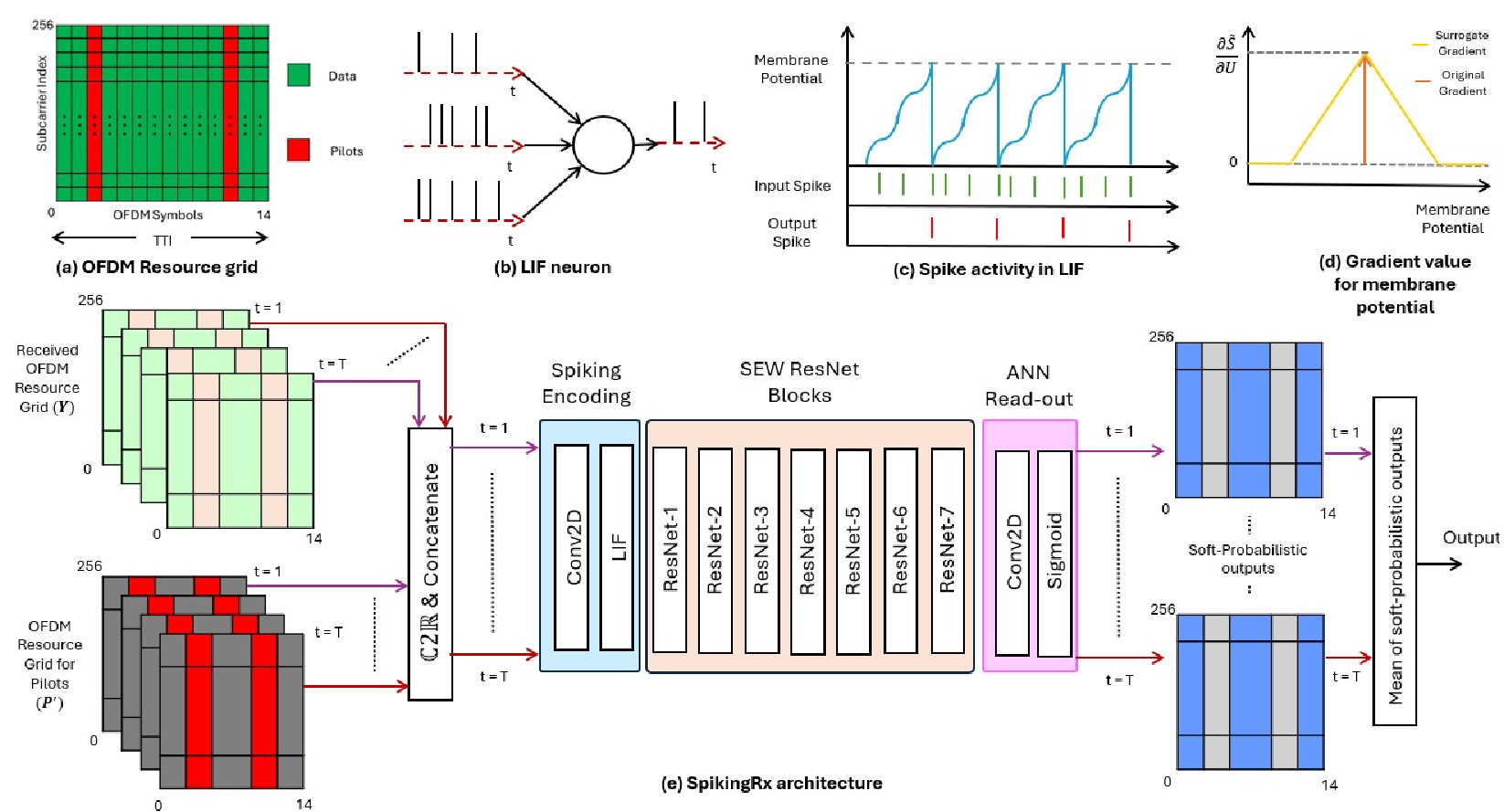}\vspace{-0.2cm}
    \caption{Illustration of NeuromorphicRx. (a) OFDM resource grid for a TTI, (b) LIF neuron with input and output spikes, (c) Spike activities in LIF neuron, (d) Gradient values concerning membrane potential, and (e) NeuromorphicRx architecture with domain-aware input signal, spiking encoding layer, 7 SEW ResNet blocks, and domain-aware output readout layer.}
    \vspace{-0.4cm}
    \label{fig:BW_SNN}
\end{figure*}

\vspace{-0.15cm}
\subsection{Spiking Neuron} 
\label{sec:spiking_neuron}
A spiking neuron operates on a weighted sum of inputs like an artificial neuron~\cite{Eshraghian2023}. The SNNs differ from ANNs in terms of neuron activations. Specifically, while the weighted sum is passed through non-linear activation functions, such as Sigmoid, Relu, and Tanh, to obtain neuron outputs in an ANN, it simply adds to the neuron's membrane potential $U(t)$ in an SNN, so that the spiking neuron fires a spike to the following neurons only if its membrane potential crosses the threshold $\theta$. Lapicque quantified the spiking neuron behaves like the Leaky Integrate-and-Fire (LIF) neuron, as shown in Fig.~\ref{fig:circuit}, which is essentially a low-pass filter circuit made up of a capacitor (C) and a resistor (R)~\cite{Eshraghian2023}. By using an RC circuit, one can model the dynamics of the passive membrane as \cite{Eshraghian2023}
\begin{align}
    \tau{dU(t)}\big/{dt} = - U\left(t\right)+I_{in}\left(t\right)R,	\label{eq:du_dt}
\end{align}
where $\tau=RC$ is the circuit time constant. Using Euler’s method, one can approximate \eqref{eq:du_dt} for discrete time as \cite{Eshraghian2023}
\begin{align}
    U\left[t\right]=\beta U\left[t-1\right]+\left(1-\beta\right)I_{in}[t],
    \label{eqn:u}
\end{align}
where $\beta=e^{-1/\tau}$ denotes the decay rate of $U[t]$. 
For simplicity, let us focus on the single input to single neuron scenario. By relaxing the physical viability constraint in \eqref{eqn:u}, the input current can be expressed as $I_{in}\left[t\right]=WX[t]$.
Considering the membrane potential reset and spiking as shown in Fig.~\ref{fig:graph}, we obtain the reset-by-subtraction expression as \cite{Eshraghian2023}
\begin{align}
    U\left[t\right]=\ \underbrace{\beta U\left[t-1\right]}_{\mathrm{decay}}+\ \underbrace{WX[t]}_{\mathrm{input}} - \underbrace{S_{out}[t-1]\theta.}_{\mathrm{reset}} \label{eq:u_updated}
\end{align}
where $S_{out}\in\{0,\ 1\}$ is a binary output spike, generated once the membrane potential exceeds the threshold $\theta$, as shown in Fig.~\ref{fig:BW_SNN}b, c, given by the shifted step function of Heaviside as

\begin{align}
    S_{out}\left[t\right]=\begin{cases}
        1, \qquad \text{if}\quad U[t]>\theta, \\
        0, \qquad \text{otherwise.}
    \end{cases}\label{eq:s_out}
\end{align}

\vspace{-0.3cm}
\subsection{NeuromorphicRx Design}
As shown in Fig.~\ref{fig:architecture_conv}, the NeuromorphicRx is employed after the removal of CP and FFT operation and replaces the signal processing blocks of channel estimation, interpolation, equalization, and symbol demapping at the receiver. 

\subsubsection{Domain-Aware Input Encoding}

SNN requires non-negative real-valued inputs to be encoded as spike trains in $C$ channels\footnote{Note that channel here refers to the number of input streams into the SNN.} and $T$ time-steps using techniques such as rate or latency/time-to-first-spike (TTFS) coding. Accordingly, a complex valued constellation with $C=1$ can be represented in terms of real-valued and normalized constellations with $C=2$. 
The encoding procedure depends on the values of $C$ and $T$ if the received (noisy) constellation symbols in \eqref{eqn:received} are directly encoded to be mapped to distinct input codewords (spike-trains).
For example, rate-coding creates $(T+1)^C$ distinct codewords encoded in at least $T \geq \left\vert\mathcal{S}_M\right\vert^{1/C}-1$ time steps, whereas latency/TTFS coding creates $T^C$ distinct codewords encoded in $T \geq \left\vert\mathcal{S}_M\right\vert^{1/C}$ time steps, where $\mathcal{S}_M$ denote a real-valued finite constellation of size $\left\vert\mathcal{S}_M\right\vert = 2^k$. Thus, time-steps will become $T\gg 2$ for the received signal in \eqref{eqn:received}. As the total energy consumption is directly proportional to $T$ in \eqref{eq:energ_cal2}, increasing $T$ leads to lower energy-efficiency. 
Furthermore, unlike scalar inputs, the modulated QAM symbols lie in a 2D complex plane. Thus, the normalization of the QAM constellation symbols (e.g., mapping $-1\rightarrow 0$, $0\rightarrow 0.5$, $1\rightarrow 1$) skews the spike-encoding. Since the negative components are suppressed, and symbols near zero are overemphasized in rate/latency coding, which leads to inaccurate spike patterns.

In order to tackle the abovementioned problems, we propose to utilize real-valued input encoding, $g(s) = (x_s, y_s) \in \mathbb{R}^2$, where I/Q values are passed directly to the NeuromorphicRx. Since $g(s)$ is injective, all symbols can be uniquely represented even with $T=1$. However, since SNNs require temporal processing, we adopt a minimal time-domain representation using $T=2$ time-steps, which is sufficient to induce spiking dynamics. We analyze impact of time-steps in Sec.~\ref{sec:timesteps_sew_trad_resnet}.

\subsubsection{Domain-Aware Input Signal}
In addition to the domain-aware input encoding described in the previous section, we leverage the domain knowledge further by setting the input signal format according to the 5G-NR frame structure.  
Let $\mathbf{Y}\in\mathbb{C}^{M\times N \times N_R}$ represent the received frequency-domain OFDM resource grid. Thus, each element of $\mathbf{Y}$ is a received modulation symbol. Apart from the received frequency-domain OFDM resource grid $\mathbf{Y}\in\mathbb{C}^{M\times N \times N_R}$, the receiver knows DMRS/pilot symbols. Thus, we provide OFDM resource grid for pilots $\mathbf{P}^\prime\in\mathbb{C}^{M\times N}$ as input in addition to the received grid $\mathbf{Y}$. The matrix $\mathbf{P}^\prime\in\mathbb{C}^{M\times N}$ contains the pilot values in corresponding pilot locations and $0$ elsewhere. We propose to provide the complete received resource grid with both the data and pilots to have a better inference about the channel in both time and frequency domains. Thus, input signal becomes $\mathbf{Q}=\mathrm{Concat}(\mathbf{Y}, \mathbf{P}^\prime)_3 \in \mathbb{C}^{M \times N \times N_R+1}$ as shown in Fig.~\ref{fig:BW_SNN}. Additional inputs, such as LS channel estimation over pilot positions and estimated noise power, can be utilized as input to the NeuromorphicRx with slightly increased complexity. Specifically, the complex input $\mathbf{Q} \in \mathbb{C}^{M \times N \times N_r+1}$ is first converted to a real-valued tensor by concatenating the real and imaginary components:
\begin{align}
    \mathbf{Q}' = \mathrm{Concat}(\text{Re}\{\mathbf{Q}\}, \text{Im}\{\mathbf{Q}\})_3 \in \mathbb{R}^{M \times N \times 2(N_R+1)}
\end{align}
To generate a time-domain representation for spiking neural networks (SNNs), $\mathbf{Q}'$ is replicated across $T$ time steps:
\begin{align}
\mathbf{Q}_{\text{in}} = \mathrm{Repeat}(\mathbf{Q}', T) \in \mathbb{R}^{M \times N \times 2(N_r+1) \times T}.
\end{align}  
\textit{Remark-1 (Domain-Aware Cheating of NeuromorphicRx during Training)} --  We propose to utilize the 2D-convolution (Conv2D) or depth-wise separable Conv-2D (DS-Conv-2D) layers so that the kernel operation is performed over the time and frequency domain to learn the channel properly. By providing the complete resource grid as input and minimizing the loss with demodulated soft bits for the whole resource grid, the NeuromorphicRx learns channel estimation and equalization from the data symbols as well as pilot symbols during the training. In contrast, 5G-NR algorithms only utilize the pilot symbols for channel estimation.

\subsubsection{NeuromorphicRx Design} The proposed NeuromorphicRx architecture consists of three main components:

\textit{(i) Spiking Encoding Layer} -- Comprises of a shared Conv-2D layer followed by LIF neuron activation. At each time step $t = 1, \dots, T$, the real-valued input $\mathbf{Q}_{\text{in}}$ is passed through shared Conv-2D layer with height $M$, width $N$, and channels $C$ that satisfies $C\gg 2(N_r\times 1)$. LIF activation is applied to each of the $M\times N\times C$ elements. Thus, we can leverage spatial encoding over $C$ channels, instead of relying on only a large number of time steps $T$. Since LIF is applied to all $C$ channels, we can have smaller $T$, avoiding the redundancy of extended temporal encoding. Thus, the network itself performs spiking conversion from raw input, eliminating the need for separate encoding mechanisms and providing different input for each time-step. Thereby, improving efficiency and performance.

\begin{figure}[t!]
    \centering
    \begin{subfigure}{0.5\textwidth}
        \centering
        \includegraphics[width=0.65\linewidth]{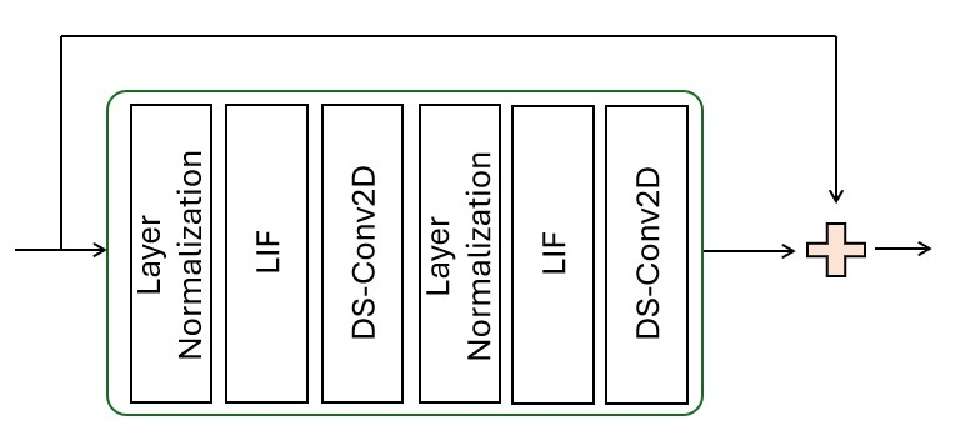}
        \caption{Traditional ResNet Block.}
        \label{fig:resnet}
    \end{subfigure}
    \begin{subfigure}{0.5\textwidth}
        \centering
        \includegraphics[width=0.6\linewidth]{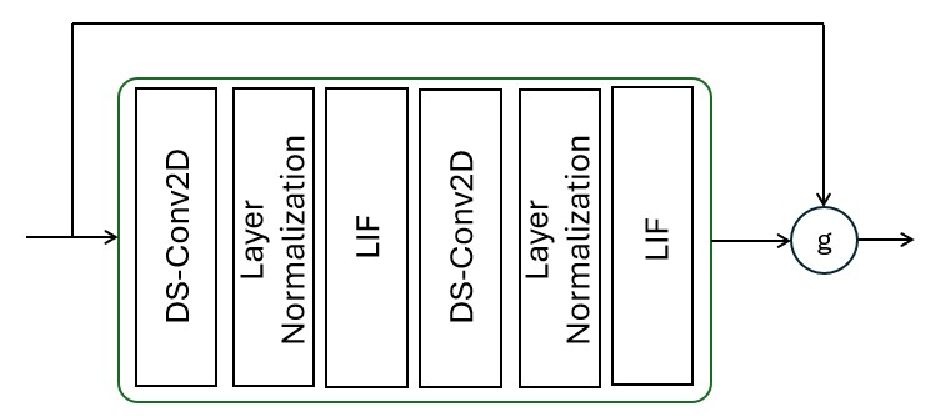}
        \caption{SEW ResNet block.}
        \label{fig:sewresnet2}
    \end{subfigure}
    \caption{Block diagrams of ResNet and SEW Resnet.}
    \vspace{-0.5cm}
\end{figure}

\textit{(ii) SEW ResNet Block Design} -- Training a deep SNN remains a challenging task due to the spiking nature of the neurons and the vanishing gradient problem, similar to the case for ANNs. 
Specifically, we can design the NeuromorphicRx with traditional ResNet blocks by modifying the ResNet blocks proposed in \cite{Honkala2021, Cammerer2020}, such that, the ReLU activation is replaced with LIF as shown in Fig.~\ref{fig:resnet}. However, such an SNN design still suffers from two major problems~\cite{Fang2021} - (1) vanishing/exploding gradient problem (even with ResNet) because gradient of spiking neuron does not satisfy $(\partial{S}/\partial{U} = 1)$ in the SGD and (2) it is unable to achieve the identity mapping because determining a firing threshold $\theta$ that ensures $U[t]>\theta$ in \eqref{eq:s_out} is challenging.
To address the abovementioned problems, we propose to utilize the SEW ResNet block~\cite{Fang2021}, as shown in Fig.~\ref{fig:sewresnet2}. Accordingly, the first problem is tackled as by having spikes at the input and output of the ResNet block. We achieve this by changing the order of the blocks from Normalization $\rightarrow$ LIF $\rightarrow$ Conv2D to Conv2D $\rightarrow$ Normalization $\rightarrow$ LIF. The second problem is tackled by utilizing the spikes' binary properties to combine the block's input and output by various logical and element-wise operations that satisfy identity mapping (detailed in Sec.~\ref{sec:i_o_sew}).  

\textit{(iii) Domain-Aware Output Readout Layer} -- As shown in Fig.~\ref{fig:BW_SNN}e, we design the NeuromorphicRx by concatenating the SNN layers (with spiking neurons) with the last ANN layer (with artificial neurons). This allows us to utilize Sigmoid activation in the last layer of the NeuromorphicRx and obtain a soft output for each bit of the detected symbol in the form of probabilities. Specifically, the logits $\widehat{a_l}\in\ \mathbb{R}$ produced in the last layer of NeuromorphicRx are passed through the Sigmoid activation function $\sigma\left(x\right) = (1+\exp{(-x)})^{-1}$ to obtain soft probabilities $\widetilde{p}(b_{m^\prime ,n}^l|\mathbf{y})$ for the $l$-th class (bits). It is shown in \cite{Cammerer2020, Gupta2021_2} that the logits correspond to the LLRs as
\begin{align}
    \widehat{a_l} = \log{\left(\frac{1-\ \widetilde{p}(b_{m^\prime ,n}^l=0|\mathbf{y})}{\ \widetilde{p}(b_{m^\prime ,n}^l=0|\mathbf{y})}\right)} = LLR_{m^\prime ,n}^l.
\end{align}
Accordingly, the soft output for $b^l_{m^{\prime},n}$ is expressed as $\widetilde{p}\left(b^l_{m^{\prime},n}=1\middle|\mathbf{y}\right)=\sigma(LLR_{m^\prime ,n}^l)$. 
NeuromorphicRx produces $T$ soft outputs or LLRs for each RE after $T$ time steps. Thus, the readout layer outputs:
\begin{align}
\widetilde{p}_t(b_{m',n}^l|\mathbf{y}) \in [0, 1]^{M' \times N}, \quad \forall t \in \{1, \dots, T\}.
\end{align}
Finally, all outputs across $T$ time steps are aggregated to produce final LLR or soft probability for each RE:
\begin{align}
\widetilde{p}(b_{m',n}^l|\mathbf{y}) = \frac{1}{T} \sum\nolimits_{t=1}^{T} \widetilde{p}_t(b_{m',n}^l|\mathbf{y}) \in [0,1]^{M' \times N}.
\end{align}
The obtained LLRs are directly used by channel decoder\footnote{It will be shown in Section~\ref{sec:training} that the soft outputs are used in the loss function for training NeuromorphicRx. However, the logits/LLRs can directly be taken as the NeuromorphicRx output to be used as input to a channel decoder after the network is deployed.}. 

\textit{Remark-2}: It is possible to design NeuromorphicRx with spiking neurons in the last layer by utilizing the rate coding. In rate coding, the predicted class is determined by the neuron that spikes most frequently. The soft-outputs $\widetilde{p}(b_{m^\prime ,n}^l|\mathbf{y})$ are obtained from the spike count by taking the mean of spikes over a total period of $T$, given as $\widetilde{p}(b_{m^\prime ,n}^l|\mathbf{y})=\sum_{t=0}^{T-1}{\vec{S_l}\left[t\right]}/T,$ where $\vec{S_l}[t]$ denote the spikes in last layer for each time step. However, in our studies, we have observed that such a design does not perform as well as the proposed design with ANN in the last layer, so we omit it for brevity.

\begin{table}[t]
\caption{The NeuromorphicRx CNN SEW ResNet.}
\label{table:4:snn_architecture}
\centering 
\renewcommand{\arraystretch}{1}
\begin{tabular}{| c | c| c | c|} 
\hline
\textbf{Layer} & \textbf{Times} & \textbf{Filter} & \textbf{Kernel} \\ [0.5ex] 
\hline\hline
Input & \multicolumn{3}{c|}{$\mathbf{Q}_{\text{in}} \in \mathbb{R}^{M \times N \times 2(N_r+1) \times T}$}\\
\hline
Conv-2D & $1$ & $128$ & $3\times 3$\\
\hline
\multicolumn{4}{|c|}{LIF}\\
\hline
Trad./SEW ResNet & $R_{sew}$  & $128$ & $3\times 3$\\
\hline
Conv-2D & $1$ & $B_t$ & $1 \times 1$\\
\hline
Output & \multicolumn{3}{c|}{LLR Values $\mathbf{LLR}\in\mathbb{R}^{M^{\prime}\times N \times T}$}\\
\hline
Sigmoid  & \multicolumn{3}{c|}{Soft outputs $\widetilde{p}(b_{m^\prime ,n}^l|\mathbf{y})\in[0, 1]^{M^{\prime}\times N \times T}$}\\
\hline
Mean  & \multicolumn{3}{c|}{Soft outputs $\widetilde{p}(b_{m^\prime ,n}^l|\mathbf{y})\in[0, 1]^{M^{\prime}\times N}$}\\
\hline
\end{tabular}
\vspace{-0.4cm}
\end{table}

The detailed architecture of NeuromorphicRx is given in Fig.~\ref{fig:BW_SNN} and Table~\ref{table:4:snn_architecture}, where $R_{sew}=7$ denotes number of ResNet blocks, until stated otherwise. We implement the NeuromorphicRx in PyTorch utilizing SNNTorch library~\cite{Eshraghian2023}.
\vspace{-0.4cm}
\subsection{Training Methodology}
\label{sec:training}
The NeuromorphicRx solves a multi-label binary classification problem by minimizing binary cross-entropy loss as 
\begin{align}
    \mathcal{L}(\cdot) &= \dfrac{1}{BMN}\sum_{l=0}^{B-1}\sum_{m=0}^{M^\prime-1}\sum_{n=0}^{N-1}b_{m^\prime ,n}^l\log\left(\widetilde{p}\left(b_{m^\prime ,n}^l\middle| \mathbf{y}\right)\right)+ \nonumber \\
    & \qquad\qquad (1-b_{m^\prime ,n}^l)\log{\left(1-\widetilde{p}\left(b_{m^\prime ,n}^l\middle| \mathbf{y}\right)\right).}
\end{align}
NNs are trained via the back-propagation method using gradients to update the weights. However, SNN suffers from the ``dead-neuron'' problem during training \cite{Eshraghian2023}. This is because the spikes are non-differentiable, such that, the gradient of the spiking neuron is zero $(\partial{S}/\partial{U} = 0)$ for all the membrane potential $(U)$ not exceeding the threshold, and $(\partial{S}/\partial{U} = \infty)$ otherwise. There are multiple methods designed to train the SNN, such as shadow training or co-learning, where an ANN is utilized to train or convert to an SNN, as done in \cite{Liu2024_2}. In this work, we propose to train the SNN from scratch without any help of a trained ANN by employing the SGD method, as shown in Fig.~\ref{fig:BW_SNN}d, which also overcomes the dead neuron problem~\cite{Friedemann2021}. Herein, the forward pass remains same as \eqref{eq:u_updated}, while during the backward pass, we approximate the non-differentiable Heaviside step-function in \eqref{eq:s_out} with a continuous differentiable function, like threshold-shifted Sigmoid function, given as $\sigma(\cdot) = (1+\exp{(\theta-U)})^{-1}$. Thus, the gradients in the backward pass are approximated as
\begin{align}
    \dfrac{\partial S}{\partial U} \rightarrow \dfrac{\partial \tilde{S}}{\partial U} = \dfrac{\exp{(\theta-U)}}{\left(\exp{(\theta-U)}+1\right)^2}.
\end{align}
Then, we can update the weights $(W)$ as
\begin{align}
    W = W - \eta \Delta_W \mathcal{L}(W),
\end{align}
where $\eta$ denotes the learning rate. In essence, the SGD enables the errors to propagate backwards irrespective of the spiking, but spiking is required to update the weights.

\textit{Remark-3:} We find that Adaptive Moment Estimation with weight decay (Adam-W) optimizer~\cite{loshchilov2019decoupledweightdecayregularization} performs better than Adam due to improved weight decay, leading to improved convergence and generalizability. 

\subsection{Domain-Aware Generalizability}
We implement the 5G-NR compliant PDSCH/PUSCH data transmission as detailed in Sec.~\ref{sec:system_model} using Nvidia's Sionna library~\cite{sionna}. We consider an OFDM resource grid of $14$ OFDM symbols and $256$ subcarriers in a single TTI. For every TTI, an arbitrary channel model is selected. Furthermore, we randomly select the RMS delay spread, Doppler shift and SNR for every channel realization. Depending on the Doppler spread, different numbers of DMRS are required, thus we consider one DMRS and two DMRS in the OFDM resource grid. The DMRS symbols span the whole frequency grid in OFDM symbols $3$ and $12$ for two-DMRS and only $3$ for one-DMRS. Randomly generated QPSK symbols form the DMRS pilot sequences. During the training of NeuromorphicRx, we focus on the domain-aware generalizability: 
\begin{itemize} 
    \item \textit{Generalizability to varying channel conditions} -- We consider the five different TDL and five different CDL channel models, each with a unique delay profile as defined by 3GPP 38.901 \cite{3GPP}. We train NeuromorphicRx on CDL-A, C, E, and TDL – A, C, E. While, we test on CDL-B, D, and TDL – B, D. 
    \item \textit{Generalizability to varying delay spread} -- We randomly sample delay spread values from $10-300$~ns. 
    \item \textit{Generalizability to varying Doppler spread} -- We randomly sample UE velocity from $0-35$~m/s.
    \item \textit{Generalizability to varying SNR} –- We randomly sample the SNR from $[0, 20]$~dB during training.
    \item \textit{Generalizability to varying DMRS} –- The model is trained for different DMRS configurations by randomly sampling one and two DMRS OFDM grids.
\end{itemize}
Please note that for generalizability under non-Gaussian noise, such as interference, we can train the NeuromorphicRx using UEs subjected to interference, by varying the signal-to-interference ratio (SIR) over a range, such as $0-20$~dB, following similar training methodology.

\begin{table}[t]
\caption{Parameters fo Training and Testing.}
\label{table:5:snn_parameters}
\centering 
\begin{tabular}{m{2.2cm}|m{2cm}|m{1.8cm}|m{1cm}} 
\hline
\textbf{Parameter} & \textbf{Training/ Validation} & \textbf{Testing} & \textbf{Random- ization} \\ [0.5ex] 
\hline
\hline
RX/TX Antennas & \multicolumn{2}{c|}{$2/1$} & None\\
\hline
Carrier Freq. & \multicolumn{2}{c|}{$4$ GHz} & None\\
\hline
Numerology & \multicolumn{2}{c|}{$1$ ($30$ kHz subcarrier spacing)} & None\\
\hline
Number of PRBs & \multicolumn{2}{c|}{$21.33$ ($256$ subcarriers)} & None\\
\hline
Symbol Duration & \multicolumn{2}{c|}{$38.02\;\mu$s} & None\\
\hline
CP Duration & \multicolumn{2}{c|}{$4.68\;\mu$s} & None\\
\hline
TTI Length & \multicolumn{2}{c|}{$14$ OFDM Symbols ($1$ ms)} & None\\
\hline
Modulation & \multicolumn{2}{c|}{16-QAM ($M_O=4$)} & None\\
\hline
Code-rate & \multicolumn{2}{c|}{0.5} & None\\
\hline
DMRS  & \multicolumn{2}{c|}{$1$ or $2$} & Uniform\\
\hline
Channel Model & CDL-A, C, E, TDL-A, C, E & CDL-B, D, TDL-B, D & Uniform\\
\hline
$E_b/N_0$ & \multicolumn{2}{c|}{$0-20$ dB} & Uniform\\
\hline
RMS Delay Spread & \multicolumn{2}{c|}{$10-300$ ns} & Uniform\\
\hline
Doppler Shift & \multicolumn{2}{c|}{$0-500$ Hz} & Uniform\\
\hline
Dataset volume & $10.24$M & $25$K & None\\
\hline
\end{tabular}
\vspace{-0.4cm}
\end{table}

\vspace{-0.4cm}
\subsection{Robustness of NeuromorphicRx}
In this section, we focus on the robustness of the NeuromorphicRx with quantized training. In our work, both ANN and SNN are trained with $32$ bits floating-point precision. However, many devices such as mobile phones, IoT devices, etc., have limited storage and processing capabilities. Further, in a wireless model update, transferring the full-precision model weights over the air will require a significant bandwidth. Thus, we focus on quantized NeuromorphicRx.

Broadly, quantized SNN can be obtained by two methods: (1) post-training-quantization (PTQ) and (2) quantization-aware training (QAT). In PTQ, a full-precision SNN is trained and then converted to lower-precision fixed-point representations. 
In QAT~\cite{nagel2022overcoming}, the quantization of weights is performed during the forward pass in training as
\begin{align}
    \widehat{\mathbf{W}} = Q(\mathbf{W}; s, l_o, u_p) = s \cdot \text{clip}\left( \left\lfloor \dfrac{\mathbf{W}}{s} \right\rceil, l_o, u_p \right),
\end{align}
where $s$ denotes the scaling factor, and $l_o$ and $u_p$ is the lower and upper quantization threshold. However, the quantization process remains non-differentiable. Thus, we ignore it during the backward pass by using the straight-through estimator, which sets the gradients to one during the backward pass, within the quantization limits, given as
\begin{align}
    \dfrac{\partial {\mathcal{L}}}{\partial \mathbf{W}} = \dfrac{\partial {\mathcal{L}}}{\partial \widehat{\mathbf{W}}} \cdot \mathbf{1}_{l_o \leq \mathbf{w}/s \leq u_p},
\end{align}
where $\mathbf{1}$ is an indicator function that gives $1$ is $l_o \leq \mathbf{W}/s \leq u_p$ and $0$, otherwise. Thus, the model learns to tackle quantization errors during the training.
Even with further calibration, the lower-precision model obtained by PTQ suffers from significant accuracy degradation~\cite{nagel2022overcoming}. Thus, we propose to employ the QAT for obtaining the quantized NeuromorphicRx. 

\begin{figure*}[t!]
    \centering
    \includegraphics[scale=0.475]{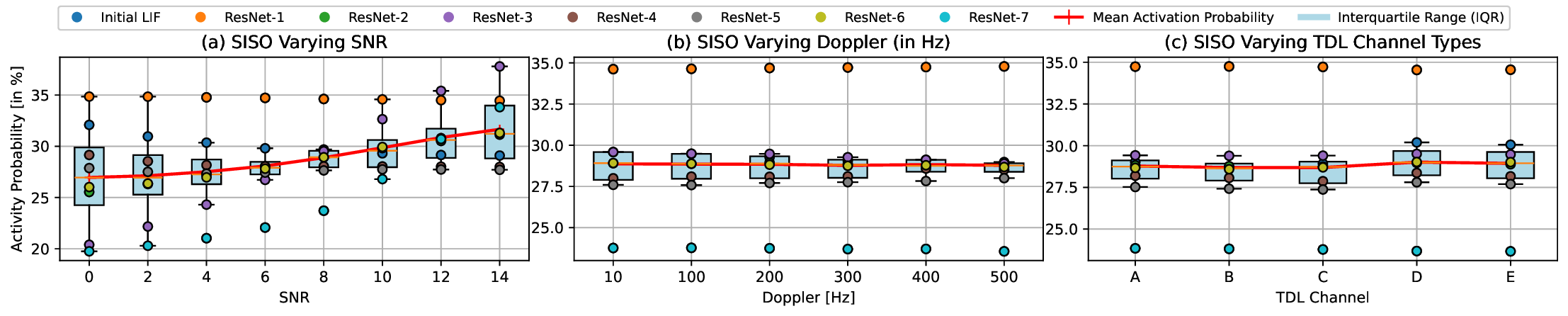}\\
    \includegraphics[scale=0.475]{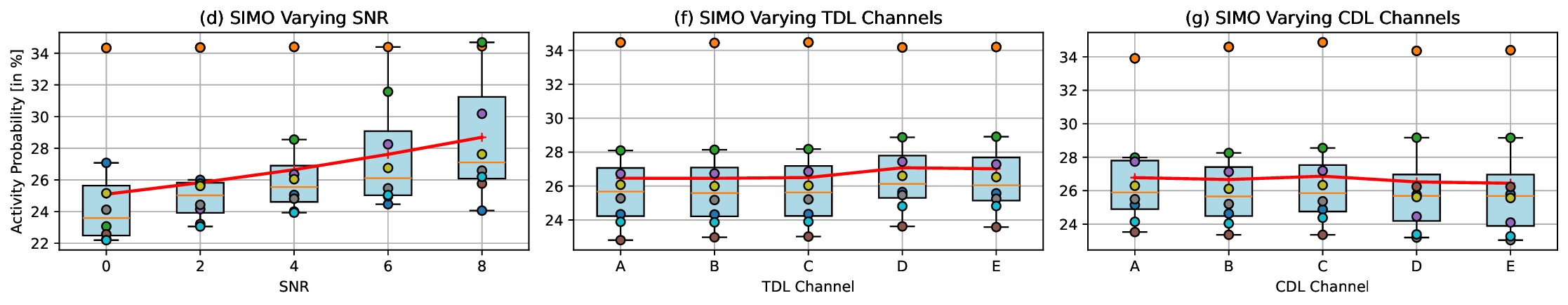}
    \caption{Activation probabilities with varying SNR $(E_b/N_0)$ for low speed under testing TDL, CDL-B, D channels, varying Doppler for fixed $E_b/N_0=8$~dB (single-antenna) and $E_b/N_0=4$~dB (multi-antenna) under testing TDL, CDL-B, D channels and all five TDL and CDL (multi-antenna) channel models in \cite{3GPP}.}
    \vspace{-0.5cm}
    \label{fig:test_activation_prob}
\end{figure*}


\vspace{-0.1cm}
\section{Ablation Study}
\label{sec:ablation_study}
In this section, we perform an ablation analysis for NeuromorphicRx to gain a more in-depth intuition of its operation. Throughout the analysis, we consider the NeuromorphicRx architecture in Table~\ref{table:4:snn_architecture} with the parameters summarized in Table~\ref{table:5:snn_parameters} and Sec.~IV B-E is performed for single antenna receivers, unless stated otherwise. Please note LDPC decoder is utilized for BLER evaluation. 

\vspace{-0.15cm}
\subsection{Neuron Activation for Varying Wireless Conditions}
The spiking neurons are said to be active when they produce discrete spikes as output at different time steps. One can calculate the spatial-temporal activation probability (in \%) of the NeuromorphicRx as~\cite{malcolm2023}
\begin{align}
    A = \frac{100a}{BTN}, \label{eq:act_prob}
\end{align}
where $a$ denotes the number of active neurons
and $N$ is the total number of neurons. 

In Fig.~\ref{fig:test_activation_prob}, we analyze the spiking activation probability (in \%) during the testing phase for single-antenna and multi-antenna scenarios with a box plot where different layers of the NeuromorphicRx are shown as scatter plot. The NeuromorphicRx obtains lower activations for initial layer/ResNet block similar to ANN-based neural networks, such as NeuralRx, where the lower activation in the initial layers is shown to gradually refine the features as data propagates through the layers [44]. We vary the SNR $(E_b/N_0)$ in  Fig.~\ref{fig:test_activation_prob}a, 5d, Doppler in Fig.~\ref{fig:test_activation_prob}b, and propagation channels in Fig.~\ref{fig:test_activation_prob}c, 5e, 5f. We can see that as the SNR is improved the activation probability also increases. A similar observation was made in~\cite{Afshar2014} for the spike time-dependent plasticity (STDP) SNN models. Although the overall activation probability across different Doppler and channel types remains similar, likely due to similar SNR, it varies subtly. Please note that neuron activation probability depends on the temporal and spatial diversity of the input. Higher activation is observed for LOS TDL and NLOS CDL due to rich multi-path with either strong delays (LOS TDL) or angular diversity (NLOS CDL). In contrast, NLOS TDL has dispersed, uncorrelated delays, and LOS CDL has limited spatial diversity, leading to lower activation. Similarly, higher activation is observed for lower Doppler shifts, possibly because slower channel variations allow more stable and consistent stimulation of neurons. 

\begin{figure}[t!]
    \centering
    \begin{subfigure}[t]{\linewidth}
        \centering
        \includegraphics[scale=0.5]{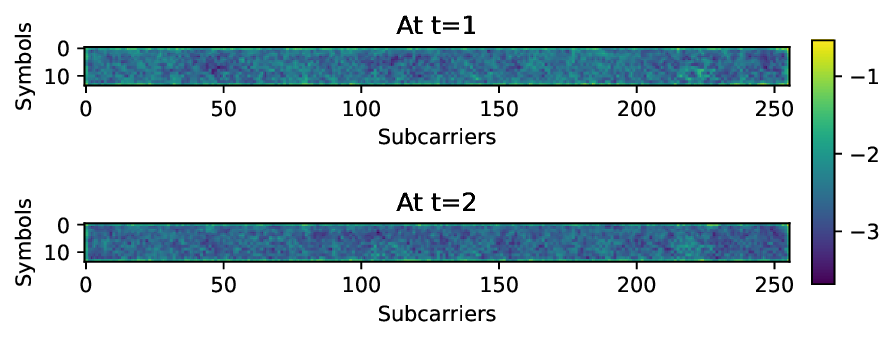}
        \caption{$E_b/N_0=0$ dB.}
        \label{fig:mem_low}
    \end{subfigure}
    \begin{subfigure}[t]{\linewidth}
        \centering
        \includegraphics[scale=0.5]{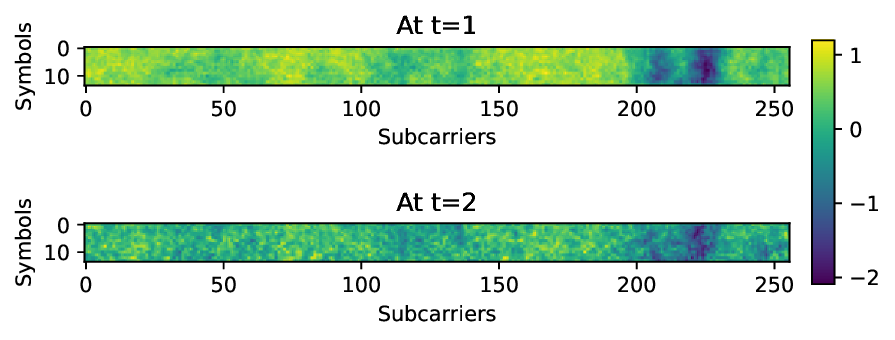}
        \caption{$E_b/N_0=14$ dB.}
        \label{fig:mem_high}
    \end{subfigure}
    \caption{Membrane potential for varying SNR for the second LIF in the SEW-ResNet-7 block.\vspace{-0.3cm}}
    \label{fig:mem_potential}
\end{figure}

To understand the above, we analyze the membrane potentials of the single-antenna NeuromorphicRx in Fig.~\ref{fig:mem_potential} for extreme SNR $(E_b/N_0)$, we note similar observations for multi-antenna scenario. Specifically, we consider the membrane potential of the second LIF of the $7^\mathrm{th}$ SEW-ResNet block because it has the most significant change in its activation probability in Fig.~\ref{fig:test_activation_prob}a. Now, let us consider \eqref{eq:du_dt}, in a noisy condition such as, low SNR, high Doppler and NLOS channels, the input current $I_{in}(t)$ observes higher fluctuations because the SNR of $I_{in}(t)$ reduces. Leading to higher variability in the membrane potential $U(t)$. As detailed in Sec.~\ref{sec:spiking_neuron}, the spiking neuron fires a spike only if its membrane potential $U(t)$ crosses the threshold $\theta$, as can be seen with \eqref{eq:u_updated} and \eqref{eq:s_out}. Thus, as seen in Fig.~\ref{fig:mem_low}, for lower SNR the $U(t)\in[-3.5, -0.5]$, which is much farther from our threshold of $\theta=1$ in the NeuromorphicRx, leading to lower activations. In contrast, for higher SNR the $U(t)\in[-2, 1]$, with most membrane potential being greater than zero, leading to higher activations. This shows that our NeuromorphicRx is passing the most relevant information in each layer to reduce energy consumption, as discussed in Sec.~\ref{sec:energy_consumpt}.

\begin{figure}
    \centering
    \begin{subfigure}[t]{\linewidth}
        \centering
        \includegraphics[scale=0.6]{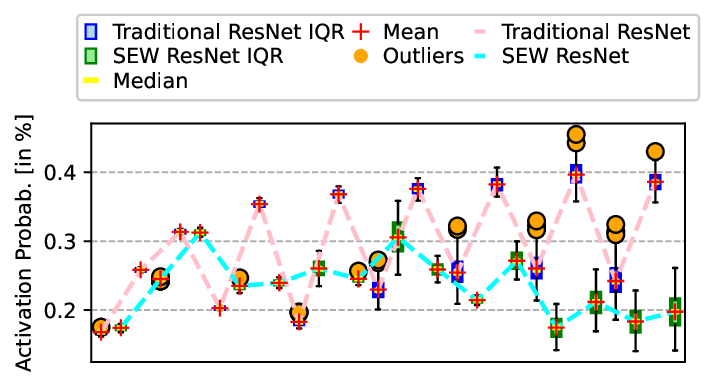}
        \includegraphics[scale=0.6]{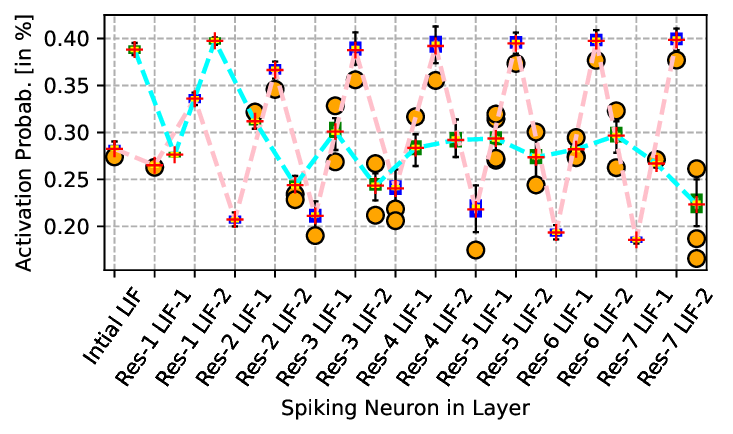}
        \caption{Activation probabilities for $T=10$ (top) and $T=2$ (bottom).}
        \label{fig:sew_res_act}
    \end{subfigure}
    \begin{subfigure}[t]{\linewidth}
        \centering
        \includegraphics[scale=0.6]{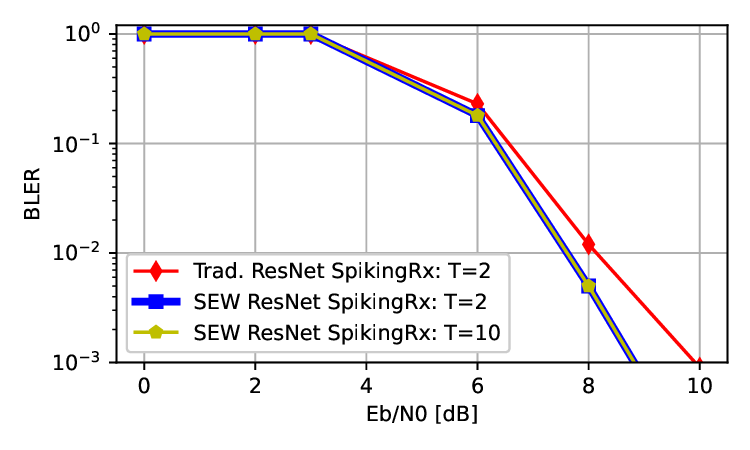}
        \caption{BLER performance.}
        \vspace{-0.1cm}
        \label{fig:sew_res_bler}
    \end{subfigure}
    \caption{SEW vs. Traditional ResNet and impact of varying time-steps.}
    \vspace{-0.4cm}
    \label{fig:sew_vs_normal_resnet}
\end{figure}

\vspace{-0.1cm}
\subsection{SEW vs. Traditional ResNet and Impact of Varying Timesteps} \label{sec:timesteps_sew_trad_resnet}
In Fig.~\ref{fig:sew_vs_normal_resnet} we analyze the traditional and SEW ResNet blocks for varying time steps $(T)$ in the spiking LIF neurons on the NeuromorphicRx. In Fig.~\ref{fig:sew_res_act}, we analyze the activation probabilities (in \%) of the NeuromorphicRx for $T=10$ and $T=2$. As one can see from the figure both traditional and SEW ResNet obtain mean activation probabilities of approximately $0.3$ for $T=2$. However, as the time step increases to $T=10$, the mean activation probabilities of the SEW and traditional ResNet become approximately $0.23$ and $0.35$, respectively. Further, the activation probability reduces with increasing time steps (T). Note that the activation probability $A$ in \eqref{eq:act_prob} is inversely proportional to $T$. We use the results in Fig.~\ref{fig:sew_vs_normal_resnet} and the methods in Sec.~\ref{sec:energy_consumpt} to calculate that SEW ResNet reduces the energy consumption by $21\%$ compared to traditional ResNet for $T=10$ and remains same for $T=2$. Note that each ResNet block has two convolutions each followed by a LIF activation. Furthermore, the second LIF activates more frequently in each ResNet block, with this contrast more prevalent with traditional ResNet blocks. Intuitively, this indicates that the second convolution layer is extracting more complex features than the first convolution layer. Moreover, the activation probability of the last few ResNet blocks for SEW ResNet is much less than that of the traditional ResNet block. Thus, traditional ResNet is gradually learning with deeper layers to extract the most influential features in the last few layers, instead of spreading the learning throughout the whole NeuromorphicRx as with the SEW ResNet block that has similar activation levels throughout. In Fig.~\ref{fig:sew_res_bler}, we analyze the BLER performance of the traditional and SEW ResNet blocks for varying $T$. The SEW ResNet outperforms the traditional ResNet, and similar performance is achieved for $T=\{2, 10\}$. This indicates that the first few time steps have the maximum `relevant' information for inference, as also shown in~\cite{kim2023exploring} by analyzing temporal Fisher information. This phenomenon occurs due to the utilization of domain-aware input encoding and signal (Sec. III-B-1,2) instead of rate or latency encoded inputs. Thus, larger time steps do not improve signal decoding performance.

\textit{Remark-4:} Although SEW ResNet was proposed to overcome the vanishing gradient problem in traditional ResNet for deep SNNs~\cite{Fang2021}, we do not observe such a phenomenon with either of these ResNet blocks in NeuromorphicRx. We find that the gradients for both of them remain similar. This occurs due to the relatively small number of blocks employed in NeuromorphicRx ({\sl e.g., $7$}) compared to the prior works using deep ResNet comprising of $50-100$ ResNet blocks~\cite{Fang2021}.


\vspace{-0.2cm}
\subsection{Input-Output Combining Operations in SEW ResNet}\label{sec:i_o_sew}
\begin{figure}[t!]
    \centering
    \includegraphics[scale=0.60]{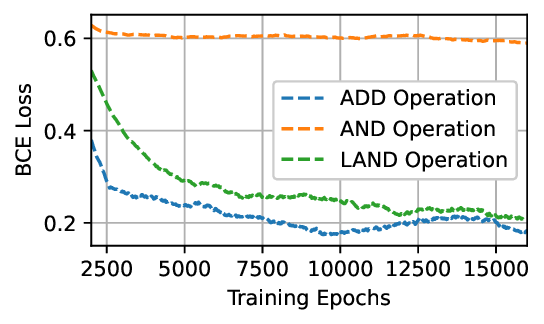}
    \caption{Operation in ResNet block.}
    \vspace{-0.4cm}
    \label{fig:various_output_oper}
\end{figure}
Since the input $(I)$ and processed signal output $(O)$ of SEW ResNet block are spiking, one can employ various logical operations to combine them as opposed to the case in traditional ResNet where a simple addition operation is used. Accordingly, one can obtain the final output of the SEW ResNet block $(g)$ using the following functions~\cite{Fang2021}:
\begin{enumerate}
    \item Addition (ADD) operation: $g=I+O$, 
    \item Logical \text{AND} operation: $g=I\ \text{AND}\ O$, 
    \item Logical IAND operation: $g=(1-I)\ \text{AND}\ O$,
\end{enumerate}
Fig.~\ref{fig:various_output_oper} shows the training loss with convergence as AND $<$ IAND $<$ ADD for 5G-NR signal demapping. Although omitted for brevity, AND requires approximately $5$ times more epochs for convergence. This is because the gradients of SEW ADD and IAND gradually increase as they move from deeper to shallower layers due to sufficient firing rates. Further, SEW ADD performs better than SEW IAND by a small margin.


\begin{figure}[t!]
    \centering
    \includegraphics[scale=0.49]{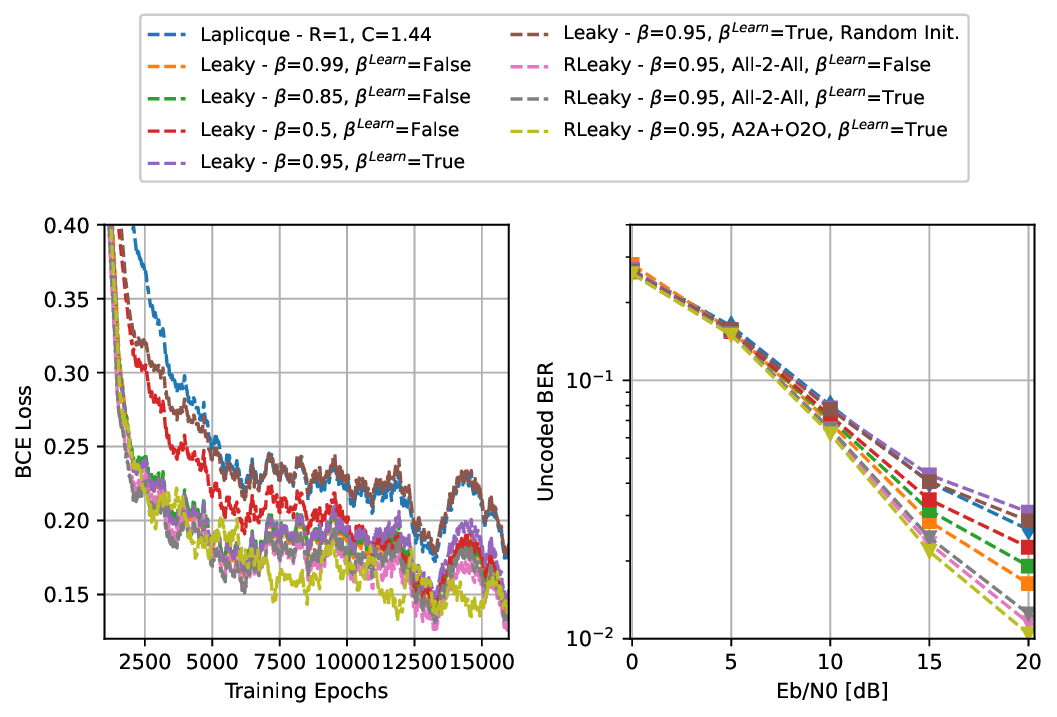}
    \caption{Varying types of spiking neuron for NeuromorphicRx.}
    \vspace{-0.3cm}
    \label{fig:diff_activation_types}
\end{figure}
\begin{figure}[t!]
    \centering
    \includegraphics[scale=0.5]{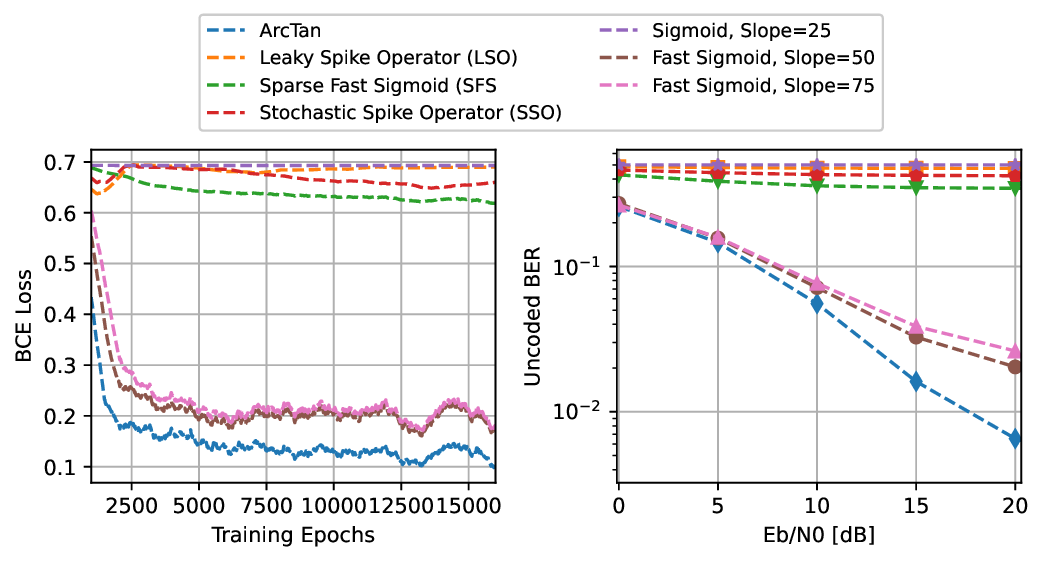}
    \caption{Varying surrogate gradients for SGD.}
    \vspace{-0.5cm}
    \label{fig:various_surrogate_gradients}
\end{figure}

\begin{figure*}[t!]
    \centering
    \includegraphics[scale=0.58]{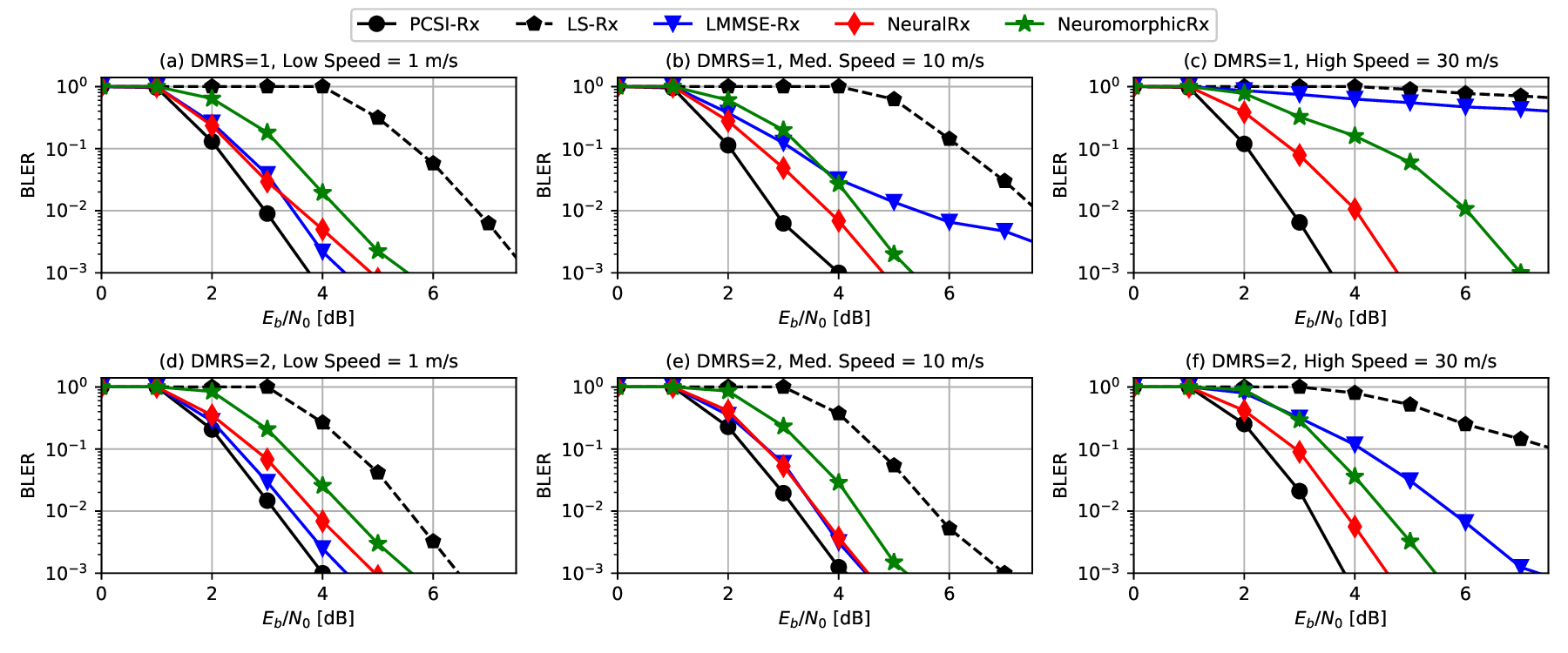}
    \caption{BLER performance of different types of receivers for varying UE speed and pilots in OFDM resource grid.}
    \vspace{-0.5cm}
    \label{fig:bler_low_med_high}
\end{figure*}

\vspace{-0.3cm}
\subsection{Spiking Neuron Activation}
In Fig.~\ref{fig:diff_activation_types}, we analyze NeuromorphicRx with varying spiking neurons, as detailed below~\cite{Eshraghian2023}:
\begin{itemize}
    \item Leaky -- As detailed in Sec.~\ref{sec:spiking_neuron}. 
    \item Lapicque -- Lapicque is qualitatively similar to the Leaky, except it requires the hyper-parameter setting of RC circuit parameters to determine decay rate $(\beta)$. 
    \item Recurrent Leaky -- Herein, the output spikes of the neuron is looped back to its input. RLeaky can be applied in two ways on the output spikes $S_{out}$ before returning back to input; (1) All-2-All recurrence, performing recurrent convolution operation, or (2) One-2-One recurrence, performing element-wise multiplication with $V$. 
\end{itemize}
We train NeuromorphicRx using Fast-Sigmoid with a slope of $25$. Firstly, Fig.~\ref{fig:diff_activation_types} shows that NeuromorphicRx with Lapicque neurons performs the worst, since the decay rate $(\beta=e^{-1/RC})$ depends on $(RC)$ hyper-parameters, requiring optimization (e.g., grid-search), which was not performed here. 

Secondly, we analyze the learnable and fixed decay rates for leaky activation. The fixed decay treats the rate as a hyperparameter, whereas the learnable decay rate treats the decay rate as a learnable parameter. Learnable decay rate performs worse due to the additional parameters, and neuron-wise variability causes unstable training, whereas, higher decay rates ($0.95$-$0.99$) enhance performance. As seen in \eqref{eqn:u}, large $\beta$ removes the input current dependency ($(1-\beta) I_{in} \rightarrow 0$), resulting in $U[t]\approx U[t-1]$, enabling NeuromorphicRx to form strong dependencies with varying time steps.

Thirdly, recurrent leaky neuron performs the best. Although there are no temporal dependencies in our input, recurrent connections stabilize the internal representations, making the NeuromorphicRx more robust to noise and wireless channel fluctuations. Overall performance in signal demapping problems ranks as Lapicque $<$ Leaky $<$ RLeaky. 

\vspace{-0.3cm}
\subsection{Surrogate Gradient Descent}
We analyze various surrogate gradients for the SGD~\cite{Eshraghian2023, Friedemann2021}: Sigmoid, Fast Sigmoid, Sparse Fast Sigmoid (SFS), Arc-tangent (ArcTan), Stochastic Spike Operator (SSO), and Leaky Spike Operator (LSO). Fig.~\ref{fig:various_surrogate_gradients} shows that only Fast Sigmoid and ArcTan enable effective learning in NeuromorphicRx. This occurs because the Fast Sigmoid gradient can better capture the sudden shifts in membrane potential than the Sigmoid gradient, due to its sharper gradients with quicker shifts around the origin. This leads to faster convergence, robustness to noise and wireless channels, while remaining more biologically plausible. Moreover, the ArcTan outperforms Fast Sigmoid significantly due to its smoother and continuous approximation to the gradients, which overcomes the gradient saturation in Fast Sigmoid, and thus, improves spike timing and increases robustness towards channel/noise impairments. Overall, performance in signal demapping problems ranks as SSO $\approx$ LSO $\approx$ SFS $\approx$ Sigmoid $<$ Fast Sigmoid $<$ ArcTan. 

\textit{Remark-5}: Note that using RLeaky instead of Leaky with ArcTan SGD provided no performance gains, despite increased complexity due to recurrent connection. Thus, the benefits of RLeaky are overshadowed by ArcTan SGD.

\section{Performance Evaluation}
\label{sec:performace_eval}
In this section, we perform further performance evaluation for SpkingRx and compare its performance with NeuralRx. Parameters are given in Table~\ref{table:5:snn_parameters} as in Section~\ref{sec:ablation_study}. We adopt the following baselines:
\begin{itemize}
    \item \textit{5G-NR - Perfect Channel State Information} (PCSI-Rx): Receiver knows the perfect CSI knowledge and performs LMMSE equalization. It acts as lower bound.
    \item \textit{5G-NR - LS Estimation} (LS-Rx): As detailed in Sec.~\ref{sec:system_model}, we consider LS channel estimation, linear channel interpolation and LMMSE equalizer. It acts as upper bound.
    \item \textit{5G-NR - LMMSE Estimation} (LMMSE-Rx): We consider LS channel estimation with LMMSE channel interpolation and LMMSE equalizer. We create frequency, time, space covariance matrices for various scenarios - single DMRS+TDL, two DMRS+TDL, single DMRS+CDL, and two DMRS+CDL channels - by varying all the channel profiles A-E, testing Doppler and RMS delay spread. It provides realistic 5G-NR performance.
    \item \textit{NeuralRx}: We implement ANN-based counterpart of NeuromorphicRx. Specifically, we replace SNN-based LIF neurons with ANN-based ReLU neurons and traditional ResNet blocks in NeuromorphicRx, with the same training and testing procedure. 
\end{itemize}

\vspace{-0.3cm}
\subsection{Evaluation of NeuromorphicRx}
In Fig.~\ref{fig:bler_low_med_high}, we evaluate the BLER performance of the NeuromorphicRx with varying speeds and pilots. Specifically, we consider three speeds: low speed ($3.6$ km/h), medium speed ($36$ km/h), and high speed ($108$ km/h), encountered in typical walking, cycling and car scenarios, respectively. In Fig.~11a-11c, we consider $1$ DMRS symbol. We can observe that only NeuralRx and NeuromorphicRx achieve BLER within $2$~dB of the lower bound of perfect CSI. Furthermore, the 5G-NR LS and LMMSE receiver is unable to decode the signal for medium-high and high speeds, as they cannot track the channel variations using a single DMRS symbol. Another important observation is that NeuromorphicRx achieves a performance around $0.5-1$~dB worse than that of NeuralRx, showing the potential of SNN in such applications. The performance degradation comes because of the higher generalization capability of NeuralRx. We note that when NeuralRx/NeuromorphicRx are trained individually for TDL/CDL channels and 1/2 DMRS the difference can reduce to $0.1$~dB. Furthermore, 5G-NR LMMSE receiver outperforms even NeuralRx for lower speed. In Fig.~11d-11f, we provide the performance results for $2$ DMRS symbols. Unlike the previous case, the 5G-NR LS receiver can decode the signal properly due to increased pilot density. Whereas, the 5G-NR LMMSE receiver can decode signal properly for all speeds. However, both LS and LMMSE still suffers from performance degradation in the high-speed scenario. On the other hand, NeuralRx and NeuromorphicRx achieve BLER performance within $1$~dB of the lower bound. 

\begin{figure}[t!]
    \centering
    \includegraphics[scale=0.5]{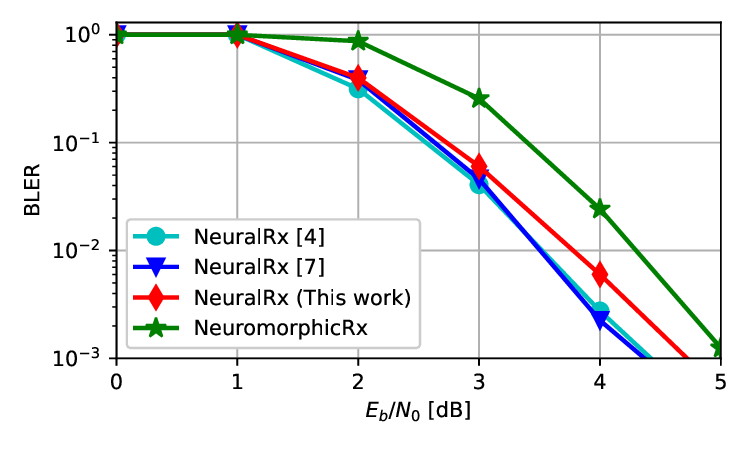}\vspace{-0.2cm}
    \caption{Comparison with SOTA NeuralRx [4], [7].}
    \vspace{-0.5cm}
    \label{fig:sota}
\end{figure}

\vspace{-0.2cm}
\subsection{Comparison with state-of-the-art (SOTA) NeuralRx}
Apart from our NeuralRx, we adopt the state-of-the-art (SOTA) NeuralRx [4] with $11$ ResNet blocks and NeuralRx [7] with $5$ ResNet blocks. In Fig.~\ref{fig:sota}, we consider two DMRS, speed varying from $1-30$ m/s, and TDL, CDL-B,D channels. Both SOTA NeuralRx [4], [7] exhibit similar performance, while the proposed NeuralRx has slightly degraded performance due to the lower complexity (as detailed in Sec.~V.D). Further, decoding performance of the NeuromorphicRx is within $1$~dB of all the SOTA and baseline NeuralRx.

\vspace{-0.2cm}
\subsection{Robustness of the NeuromorphicRx}
\begin{figure}[t!]
    \centering
        \includegraphics[scale=0.6]{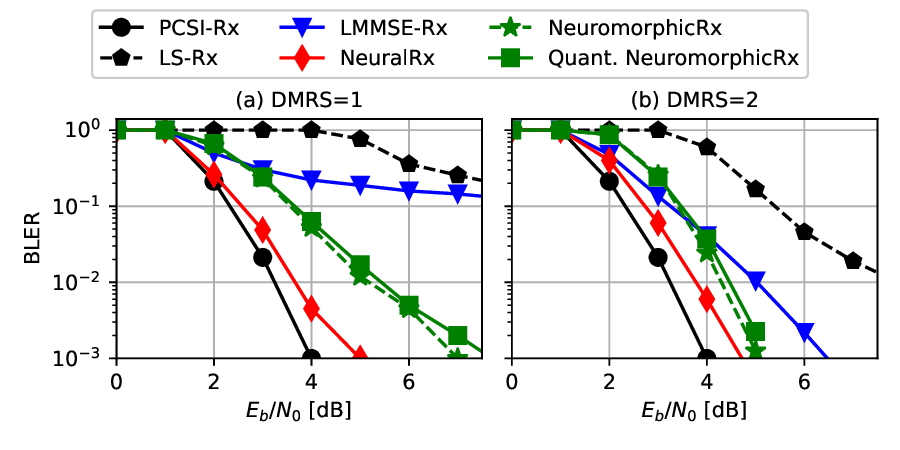}\vspace{-0.2cm}
    \caption{Robustness of NeuromorphicRx.}
    \vspace{-0.4cm}
    \label{fig:robustSNN}
\end{figure}
In Fig.~\ref{fig:robustSNN}, we analyze the robustness of the NeuromorphicRx. The results show that NeuromorphicRx with $8$-bit quantized weights achieves only $0.1$~dB worse performance than that of full-precision NeuromorphicRx with $32$ bits. Accordingly, one can conclude that NeuromorphicRx with quantize-aware training (QAT) achieves a significant complexity gain with a negligible error rate performance degradation.

\vspace{-0.3cm}
\subsection{Energy Consumption}
\label{sec:energy_consumpt}
Finally, we compare the computational energy cost for the NeuromorphicRx and NeuralRx. The total number of floating point operations (FLOPS) determines the entire processing overhead, which roughly corresponds to the quantity of Matrix-Vector Multiplications~\cite{Xie2022}. Thus, the number of FLOPS for the $l$-th layer in the ANN for $l=\{1, ..., L\}$ as 
\begin{align}
&\text{FLOPS}_{\text{NeuralRx}}(l)\hspace{-0.1cm} = \nonumber \\\hspace{-0.1cm} 
    &\begin{cases}
         K_{\text{eff}}^2 C_{in}  H_{out}  W_{out}  C_{out}, \  \text{if}\; l :=\;\text{Conv2D}, \nonumber\\
         D_m H_{out}  W_{out} C_{in} (K_{\text{eff}}^2+C_{out}), \  \text{if}\; l :=\;\text{DS-Conv2D}, \nonumber\\
        C_{in}  H_{out}  W_{out},
        \  \text{if}\; l :=\;\text{Normalization}, 
    \end{cases}
\end{align}
where, $H_{out}$ and $W_{out}$ represent the height and width of the output, $C_{in}$ and $C_{out}$ indicate the input and output channel, $K_{\text{eff}}=D_l(K_l-1)+1$ denotes the effective kernel size based on the kernel size $K_l$ and dilation rate $D_l$, and $D_m$ is the depth multiplier in DS-Conv2D. For the $l$-th layer in the SNN, the average firing rate per neuron (spiking rate) is given as
\begin{align}
    R_s(l) = {\sum\nolimits_{t=1}^T N_s^t(l)}\big/{N_n(l)}, \quad \forall\;l<L \label{eq:energ_cal2}
\end{align}
where $N_s^t$ and $N_n$ denote the number of spikes of the $l$-th layer for all the time steps $T$ and number of spiking neurons in that layer, respectively. Note that real-valued inputs are passed as input to Conv-2D layer in spiking encoding (Sec.~III.B-3-i) and the last layer-L readout layer (Sec.~III.B-3-iii) utilizes sigmoid activation. For the SNN-based NeuromorphicRx, the FLOP only happens if a spike is emitted. Thus, FLOP count for the $l$-th layer in the SNN, where $l>1$, is given as 
\begin{align}
    \text{FLOPS}_{\text{NeuromorphicRx}}(l)\!=\!\text{FLOPS}_{\text{NeuralRx}}(l) R_s(l-1)
\end{align}
\begin{table}[t!]
\caption{Energy consumption for varying operations~\cite{Horowitz2014}.}
\label{table:3:energy_consump_values}
\centering 
\renewcommand{\arraystretch}{1}
\begin{tabular}{| c | c |} 
 \hline
 \textbf{Operation} & \textbf{Energy (pJ)} \\ [0.5ex] 
 \hline\hline
 32 bit FP MULT $(E_{MULT})$ & 3.7 \\
 32 bit FP ADD $(E_{ADD})$ & 0.9 \\
 32 bit FP MAC $(E_{MAC} = E_{MULT}+E_{ADD})$ & 4.6 \\
 32 bit FP AC $(E_{AC})$ & 0.9 \\
 8 bit FP MAC $(E_{MAC})$ & 1.1 \\
 8 bit FP AC $(E_{AC})$ & 0.2 \\
 \hline
\end{tabular}\vspace{-0.25cm}
\end{table}
\begin{table}[t!]
\caption{Comparison of parameters, FLOPs, and energy consumption for 5G-NR-Rx, NeuralRx, and NeuromorphicRx.}
\label{table:energy_consump_results}
\centering
\renewcommand{\arraystretch}{1.2}
\begin{tabular}{|l|r|r|r|}
\hline
\textbf{Algorithm} & \textbf{Params (K)} & \textbf{FLOPs (GFLOPs)} & \textbf{Energy (nJ)} \\
\hline \hline
\rowcolor{gray!10} \multicolumn{4}{|l|}{\textbf{Traditional 5G-NR Receivers}} \\
PCSI-Rx            & 0       & 0.00051    & 2.50 \\
LS-Rx              & 0       & 0.00055    & 2.71 \\
LMMSE-Rx           & 0       & 3410.66    & 16712.24 \\
\hline
\rowcolor{gray!10} \multicolumn{4}{|l|}{\textbf{ANN-based NeuralRx (Varying Depth)}} \\
5 ResNets          & 190.34  & 1.36       & 6671.79 \\
7 ResNets          & 262.53  & 1.88       & 9202.91 \\
9 ResNets          & 334.72  & 2.39       & 11734.03 \\
11 ResNets         & 406.92  & 2.91       & 14265.15 \\
\hline
\rowcolor{gray!10} \multicolumn{4}{|l|}{\textbf{SNN-based NeuromorphicRx (Varying Depth)}} \\
5 ResNets          & 190.34  & 1.15       & 1003.33 \\
7 ResNets          & 262.53  & 1.51       & 1207.20 \\
9 ResNets          & 334.72  & 1.77       & 1326.26 \\
11 ResNets         & 406.92  & 2.07       & 1474.09 \\
\hline
\rowcolor{gray!10} \multicolumn{4}{|l|}{\textbf{Other ANN-based NeuralRx Implementations}} \\
Ref [4]            & 604.23  & 4.99       & 24480.66 \\
Ref [7]            & 185.99  & 2.38       & 11667.72 \\
Ours               & 262.53  & 1.88       & 9202.91 \\
\hline
\rowcolor{gray!10} \multicolumn{4}{|l|}{\textbf{SNN-based NeuromorphicRx (Time Variation, Quantization)}} \\
$T=10$             & 262.53  & 2.22       & 5112.72 \\
$T=2$              & 262.53  & 1.51       & 1207.20 \\
Quant. $T=2$       & 262.53  & 1.46       & 260.62 \\
\hline
\end{tabular}
\vspace{-0.5cm}
\end{table}
\begin{figure*}[t!]
    \centering
        \includegraphics[scale=0.45]{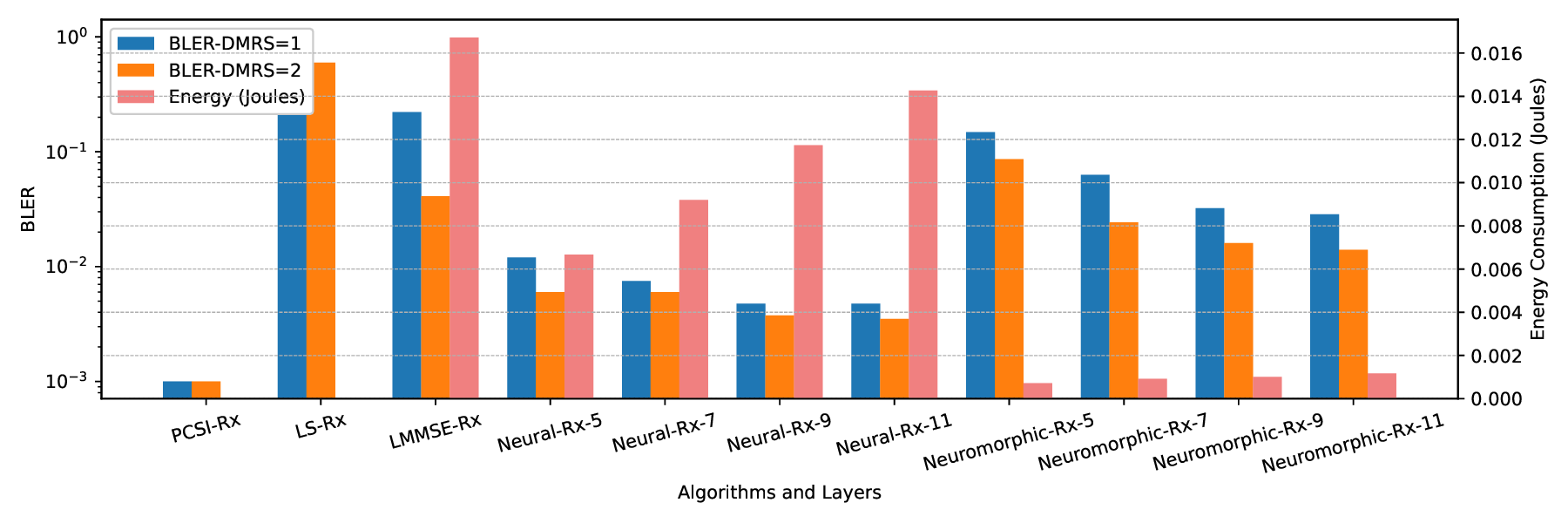}\vspace{-0.2cm}
    \caption{BLER performance and energy consumption comparison at SNR=$4$~dB for varying network sizes.}\vspace{-0.35cm}
\end{figure*}
For each input, the summation of a neuron's weighted inputs in an ANN requires a MAC operation, while spikes in SNNs require an AC operation. In NeuromorphicRx, the first Conv2D layer performs MAC operations on real-valued inputs, while the final Conv2D readout layer, preceded by LIF activation, uses AC operations. The energy cost of the sigmoid activation is computed separately due to its nonlinear nature. Thus, the energy consumed by the $l$-th layer is given by:
\begin{align}
    E_{\text{NeuralRx}}(l) &= \text{FLOPS}_{\text{NeuralRx}}(l) E_{MAC}, \\
    E_{\text{NeuromorphicRx}}(l) &= 
    \begin{cases}
        \text{FLOPS}_{\text{NeuralRx}}(l) E_{MAC}, \; \forall\;l=1 \nonumber\\
        \text{FLOPS}_{\text{NeuromorphicRx}}(l) E_{AC}, \; \text{otherwise} \nonumber
    \end{cases} 
\end{align}
Considering $45$nm CMOS technology, we can calculate the $E_{MAC}$ and $E_{AC}$ as in Table~\ref{table:3:energy_consump_values}~\cite{Horowitz2014}. For $Q$-bit quantization, we consider $E_{MAC}\propto Q^{1.25}$ and $E_{AC}\propto Q$~\cite{Datta2021}. In Table~\ref{table:energy_consump_results}, we evaluate the number of parameters, FLOPs, and energy consumption for 5G-NR algorithms, NeuralRx, and NeuromorphicRx across architectures with $5$, $7$, $9$, and $11$ ResNet blocks (Table~I). Parameters indicate model size, FLOPs reflect computational complexity, and energy consumption measures practical efficiency. Unlike NeuralRx and NeuromorphicRx, 5G-NR algorithms require no parameter storage, saving memory. LS-Rx consumes $482\times$ less energy than NeuromorphicRx, however, its performance is significantly worse. LMMSE-Rx achieves better decoding but uses $13.81\times$ more energy. As ResNet depth increases, NeuromorphicRx achieves higher energy savings over NeuralRx (from $6.7\times$ to $9.7\times$) due to efficient sparse activations in larger networks. It also reduces energy by $20.3\times$ and $9.7\times$ compared to [4], [7], with minor performance loss (Fig.~12). In Fig.~14, the BLER and energy comparison at SNR = 4 dB show NeuralRx gains~$0.3$~dB beyond $7$ blocks, and NeuromorphicRx shows negligible improvement after $9$ blocks. Thus, $7$ ResNet blocks offer the best performance-energy trade-off. In summary, NeuromorphicRx achieves $1.8\times$ and $7.6\times$ reduced energy consumption compared to NeuralRx for $T=10$ and $T=2$ time steps, respectively. Further, quantized NeuromorphicRx can provide an additional $4.6\times$ reduced energy consumption.

\vspace{-0.35cm}
\section{Conclusion and Future Works}
In this work, we proposed an SNN-based NeuromorphicRx for any 5G-NR-compliant OFDM system. Instead of spike encoding, we utilize the real-valued OFDM resource grid with pilot signals as input. NeuromorphicRx is designed using deep CNN and SEW ResNet blocks concatenated with an artificial neuron layer for obtaining soft-output or LLR values, ensuring compatibility to channel decoders. We utilize SGD for training and focus on the generalizability across varying 3GPP TDL/CDL channels, SNR, Doppler/delay spread, and DMRS configurations. We proposed quantization-aware training to improve robustness for IoT deployment. Specifically focusing on 5G-NR signal decoding, we improve interpretability of NeuromorphicRx by analyzing the spiking activations and membrane potentials and performed ablation studies to determine the optimal neurons, time-steps, surrogate functions, and SEW ResNet combining operations. Extensive analysis shows NeuromorphicRx achieves BLER within $1.2$~dB of an LMMSE receiver with perfect CSI and matches ANN-based baselines with up to $7.6\times$ lower energy, further improved by $4.6\times$ with QAT.
Possible energy-efficient future works include:
\begin{itemize}
    \item \textit{Calibration-based QAT} -- Quantization robustness can be possibly improved for deployment on fixed-point neuromorphic platforms~\cite{Hyeryung2021, Skatchkovsky2021_2, Rajendran2019} by adjusting the scaling factors and thresholds (without making them learnable) by utilizing training-time statistics (e.g., weight ranges, percentile clipping).
    \item \textit{Neuromorphic Joint Source-Channel Coding (JSCC)} -- Spike-encoding (rate/latency) remains ideal for source data. Designing an E2E neuromorphic JSCC (extending ANN-based~\cite{Jialong2023}) by training SNN-based encoders and decoders that performs both compression and error protection. The encoder maps the input spikes to binary values required by the channel coding, making SNN ideal for JSCC, and enabling low-latency E2E communications. 
    \item \textit{Neuromorphic Channel Decoder} -- Take advantage of binary and event-driven nature of spiking neurons for modeling the path metrics and message passing algorithms in the classical decoders, e.g., Viterbi, LDPC (extending ANN-based~\cite{Tobias2017}), allowing parallel and low-latency decoding.
    \item \textit{E2E Neuromorphic Transceiver for Pilotless Communications} -- Replace symbol mapping block with a SNN at the transmitter and performing E2E training for custom modulation design for pilotless transmission, improving the throughput (extending ANN-based~\cite{Faycal2022}). Transmitter-side SNN presents challenge of maintaining constellation structure for the time-step-dependent outputs that cannot be averaged directly like the NeuromorphicRx.
    \item \textit{Bio-Inspired Supervised-STDP (SSTDP) Training} -- STDP is an unsupervised biologically-inspired training method. SSTDP~\cite{liu2021sstdp} can be utilized for NeuromorphicRx to combine the local STDP-based learning in early layers with SGD-based learning in deeper layers. Our initial experiments show that it requires significant computational challenges due to the continuous monitoring of spike trace, per-time-step updates and lack of GPU implementation frameworks. Although, SSTDP performance remains promising, it still trails SGD performance.
\end{itemize}

\vspace{-0.15cm}
\renewcommand{\refname}{REFERENCES}
\bibliographystyle{IEEEtran}
\bibliography{mainbib}

@article{Dorner2018,
  author={Dörner, Sebastian and Cammerer, Sebastian and Hoydis, Jakob and Brink, Stephan ten},
  journal={IEEE J. Sel. Top. Signal Process. }, 
  title={Deep Learning Based Communication Over the Air}, 
  year={2018},
  volume={12},
  number={1},
  pages={132-143},
  doi={10.1109/JSTSP.2017.2784180}}

@ARTICLE{Hoydis2021,
  author={Hoydis, Jakob and Aoudia, Fayçal Ait and Valcarce, Alvaro and Viswanathan, Harish},
  journal={IEEE Commun. Mag.}, 
  title={Toward a {6G AI}-Native Air Interface}, 
  year={2021},
  volume={59},
  number={5},
  pages={76-81},
  doi={10.1109/MCOM.001.2001187}}

@ARTICLE{Honkala2021,
  author={Honkala, Mikko and Korpi, Dani and Huttunen, Janne M. J.},
  journal={IEEE Trans. Wireless Commun.}, 
  title={{DeepRx}: Fully Convolutional Deep Learning Receiver}, 
  year={2021},
  volume={20},
  number={6},
  pages={3925-3940},
  doi={10.1109/TWC.2021.3054520}}

@INPROCEEDINGS{Gupta2023,
  author={Gupta, Ankit and Bishnu, Abhijeet and Ratnarajah, Tharmalingam and Adeel, Ahsan and Hussain, Amir and Sellathurai, Mathini},
  booktitle={GLOBECOM 2023 - 2023 IEEE Global Communications Conference}, 
  title={Deep Learning-Based Receiver Design for {IoT} Multi-User Uplink {5G-NR} System}, 
  year={2023},
  volume={},
  number={},
  pages={4110-4115},
  doi={10.1109/GLOBECOM54140.2023.10437776}}

@ARTICLE{Faycal2022,
  author={Ait Aoudia, Fayçal and Hoydis, Jakob},
  journal={IEEE Trans. Wireless Commun.}, 
  title={End-to-End Learning for {OFDM}: From Neural Receivers to Pilotless Communication}, 
  year={2022},
  volume={21},
  number={2},
  pages={1049-1063},
  doi={10.1109/TWC.2021.3101364}}

@ARTICLE{Pihlajasalo2023,
  author={Pihlajasalo, Jaakko and Korpi, Dani and Honkala, Mikko and Huttunen, Janne M. J. and Riihonen, Taneli and Talvitie, Jukka and Brihuega, Alberto and Uusitalo, Mikko A. and Valkama, Mikko},
  journal={IEEE Trans. Wireless Commun.}, 
  title={Deep Learning {OFDM} Receivers for Improved Power Efficiency and Coverage}, 
  year={2023},
  volume={22},
  number={8},
  pages={5518-5535},
  doi={10.1109/TWC.2023.3235059}}

@ARTICLE{Raviv2023,
  author={Raviv, Tomer and Park, Sangwoo and Simeone, Osvaldo and Eldar, Yonina C. and Shlezinger, Nir},
  journal={IEEE Trans. Wireless Commun.}, 
  title={Online Meta-Learning for Hybrid Model-Based Deep Receivers}, 
  year={2023},
  volume={22},
  number={10},
  pages={6415-6431},
  doi={10.1109/TWC.2023.3241841}}

@ARTICLE{Xie2024,
  author={Xie, Yihang and Teh, Kah Chan and Kot, Alex C.},
  journal={IEEE Trans. Commun.}, 
  title={Comm-Transformer: A Robust Deep Learning-Based Receiver for {OFDM} System Under {TDL} Channel}, 
  year={2024},
  volume={72},
  number={4},
  pages={2014-2026},
  doi={10.1109/TCOMM.2023.3343787}}

@ARTICLE{Mei2024,
  author={Mei, Ruru and Wang, Zhugang and Chen, Xuan},
  journal={IEEE Trans. Cogn. Commun. Netw.}, 
  title={{CRNN-ResNet}: Combined {CRNN} and {ResNet} Networks for {OFDM} Receivers}, 
  year={2024},
  volume={},
  number={},
  pages={1-1},
  doi={10.1109/TCCN.2024.3378225}}

@ARTICLE{Sun2024,
  author={Sun, Yi and Shen, Hong and Li, Bingqing and Xu, Wei and Zhu, Pengcheng and Hu, Nan and Zhao, Chunming},
  journal={IEEE Trans. Wireless Commun.}, 
  title={Trainable Joint Channel Estimation, Detection and Decoding for {MIMO} {URLLC} Systems}, 
  year={2024},
  volume={},
  number={},
  pages={1-1},
  doi={10.1109/TWC.2024.3388603}}

@INPROCEEDINGS{Korpi2023,
  author={Korpi, Dani and Honkala, Mikko and Huttunen, Janne M.J.},
  booktitle={2023 57th Asilomar Conference on Signals, Systems, and Computers}, 
  title={Deep Learning-Based Pilotless Spatial Multiplexing}, 
  year={2023},
  volume={},
  number={},
  pages={1025-1029},
  doi={10.1109/IEEECONF59524.2023.10477093}}

@article{Wan23,
  title={Graph Neural Network-based Joint Equalization and Decoding},
  author={Clausius, Jannis and Geiselhart, Marvin and Tandler, Daniel and Brink, Stephan ten},
  journal={arXiv preprint arXiv:2401.16187},
  year={2024}}

@ARTICLE{Hyeryung2021,
  author={Jang, Hyeryung and Skatchkovsky, Nicolas and Simeone, Osvaldo},
  journal={IEEE Commun. Lett.}, 
  title={Spiking Neural Networks—Part {I}: Detecting Spatial Patterns}, 
  year={2021},
  volume={25},
  number={6},
  pages={1736-1740},
  doi={10.1109/LCOMM.2021.3050207}}

@ARTICLE{Skatchkovsky2021_2,
  author={Skatchkovsky, Nicolas and Jang, Hyeryung and Simeone, Osvaldo},
  journal={IEEE Commun. Lett.}, 
  title={Spiking Neural Networks—Part {III}: Neuromorphic Communications}, 
  year={2021},
  volume={25},
  number={6},
  pages={1746-1750},
  doi={10.1109/LCOMM.2021.3050212}}

@ARTICLE{Rajendran2019,
  author={Rajendran, Bipin and Sebastian, Abu and Schmuker, Michael and Srinivasa, Narayan and Eleftheriou, Evangelos},
  journal={IEEE Signal Process. Mag.}, 
  title={Low-Power Neuromorphic Hardware for Signal Processing Applications: A Review of Architectural and System-Level Design Approaches}, 
  year={2019},
  volume={36},
  number={6},
  pages={97-110},
  doi={10.1109/MSP.2019.2933719}}

@ARTICLE{Ge2023,
  author={Ge, Xiaokai and Hu, Xianzhi and Dai, Xuchu},
  journal={IEEE Commun. Lett.}, 
  title={Unsupervised Learning Feature Estimation for {MISO} Beamforming by Using Spiking Neural Networks}, 
  year={2023},
  volume={27},
  number={4},
  pages={1165-1169},
  doi={10.1109/LCOMM.2023.3246052}}

@ARTICLE{Ortiz2024,
  author={Ortiz, Flor and Skatchkovsky, Nicolas and Lagunas, Eva and Martins, Wallace A. and Eappen, Geoffrey and Daoud, Saed and Simeone, Osvaldo and Rajendran, Bipin and Chatzinotas, Symeon},
  journal={IEEE Trans. Mach. Learn. Commun. Netw.}, 
  title={Energy-Efficient on-Board Radio Resource Management for Satellite Communications via Neuromorphic Computing}, 
  year={2024},
  volume={2},
  number={},
  pages={169-189},
  doi={10.1109/TMLCN.2024.3352569}}

@ARTICLE{Vogginger2022,
AUTHOR={Vogginger, Bernhard  and Kreutz, Felix  and López-Randulfe, Javier  and Liu, Chen  and Dietrich, Robin  and Gonzalez, Hector A.  and Scholz, Daniel  and Reeb, Nico  and Auge, Daniel  and Hille, Julian  and Arsalan, Muhammad  and Mirus, Florian  and Grassmann, Cyprian  and Knoll, Alois  and Mayr, Christian},
TITLE={Automotive Radar Processing With Spiking Neural Networks: Concepts and Challenges},
JOURNAL={Frontiers in Neuroscience},
VOLUME={16},
YEAR={2022},
DOI={10.3389/fnins.2022.851774},
ISSN={1662-453X}}

@ARTICLE{Liu2024,
  author={Liu, Shiya and Mohammadi, Nima and Yi, Yang},
  journal={IEEE Trans. Green Commun. Netw.}, 
  title={Quantization-Aware Training of Spiking Neural Networks for Energy-Efficient Spectrum Sensing on Loihi Chip}, 
  year={2024},
  volume={8},
  number={2},
  pages={827-838},
  doi={10.1109/TGCN.2023.3337748}}

@ARTICLE{Liu2024_2,
  author={Liu, Shiya and Liang, Yibin and Yi, Yang},
  journal={IEEE Trans. Sustainable Comput.}, 
  title={{DNN-SNN} Co-Learning for Sustainable Symbol Detection in {5G} Systems on Loihi Chip}, 
  year={2024},
  volume={9},
  number={2},
  pages={170-181},
  doi={10.1109/TSUSC.2023.3324339}}

@ARTICLE{Chen2023,
  author={Chen, Jiechen and Skatchkovsky, Nicolas and Simeone, Osvaldo},
  journal={IEEE Trans. Cogn. Commun. Netw.}, 
  title={Neuromorphic Wireless Cognition: Event-Driven Semantic Communications for Remote Inference}, 
  year={2023},
  volume={9},
  number={2},
  pages={252-265},
  doi={10.1109/TCCN.2023.3236940}}

@article{Chen2024,
  title={Neuromorphic Split Computing with Wake-Up Radios: Architecture and Design via Digital Twinning},
  author={Chen, Jiechen and Park, Sangwoo and Popovski, Petar and Poor, H. Vincent and Simeone, Osvaldo},
  journal={arXiv preprint arXiv:2404.01815},
  year={2024}
}

@INPROCEEDINGS{Borsos2022,
  author={Borsos, Tamás and Condoluci, Massimo and Daoutis, Marios and Hága, Péter and Veres, András},
  booktitle={2022 IEEE Wireless Communications and Networking Conference}, 
  title={Resilience Analysis of Distributed Wireless Spiking Neural Networks}, 
  year={2022},
  volume={},
  number={},
  pages={2375-2380},
  doi={10.1109/WCNC51071.2022.9771543}}

@Article{Velusamy2023,
AUTHOR = {Velusamy, Gandhimathi and Lent, Ricardo},
TITLE = {Delay-Packet-Loss-Optimized Distributed Routing Using Spiking Neural Network in Delay-Tolerant Networking},
JOURNAL = {Sensors},
VOLUME = {23},
YEAR = {2023},
NUMBER = {1},
ARTICLE-NUMBER = {310},
PubMedID = {36616907},
ISSN = {1424-8220},
DOI = {10.3390/s23010310}}

@ARTICLE{Wen2024,
  author={Wen, Dingzhu and Liu, Peixi and Zhu, Guangxu and Shi, Yuanming and Xu, Jie and Eldar, Yonina C. and Cui, Shuguang},
  journal={IEEE Trans. Wireless Commun.}, 
  title={Task-Oriented Sensing, Computation, and Communication Integration for Multi-Device Edge {AI}}, 
  year={2024},
  volume={23},
  number={3},
  pages={2486-2502},
  doi={10.1109/TWC.2023.3303232}}

@ARTICLE{Hamedani2021,
  author={Hamedani, Kian and Liu, Lingjia and Yi, Yang},
  journal={IEEE Trans. Green Commun. Netw.}, 
  title={Energy Efficient {MIMO-OFDM} Spectrum Sensing Using Deep Stacked Spiking Delayed Feedback Reservoir Computing}, 
  year={2021},
  volume={5},
  pages={484-496},
  doi={10.1109/TGCN.2020.3046725}}

@article{Hamedani2020, 
title={Deep Spiking Delayed Feedback Reservoirs and Its Application in Spectrum Sensing of {MIMO-OFDM} Dynamic Spectrum Sharing}, 
journal={Proceedings of the AAAI Conference on Artificial Intelligence}, 
volume={34}, 
DOI={10.1609/aaai.v34i02.5484}, 
author={Hamedani, Kian and Liu, Lingjia and Liu, Shiya and He, Haibo and Yi, Yang}, 
year={2020}, 
month={Apr.}, 
pages={1292-1299} }

@ARTICLE{Dakic2024,
  author={Dakic, Kosta and Al Homssi, Bassel and Walia, Sumeet and Al-Hourani, Akram},
  journal={IEEE Trans. Aerosp. Electron. Syst.}, 
  title={Spiking Neural Networks for Detecting Satellite Internet of Things Signals}, 
  year={2024},
  volume={60},
  number={1},
  pages={1224-1238},
  doi={10.1109/TAES.2023.3334216}}

@ARTICLE{Gupta2022,
  author={Gupta, Ankit and Sellathurai, Mathini},
  journal={IEEE Trans. Cogn. Commun. Netw.}, 
  title={A Novel Average Autoencoder-Based Amplify-and-Forward Relay Networks With Hardware Impairments}, 
  year={2022},
  volume={8},
  number={2},
  pages={615-630},
  doi={10.1109/TCCN.2022.3164901}}

@ARTICLE{Gupta2023_2,
  author={Gupta, Ankit and Sellathurai, Mathini and Ratnarajah, Tharmalingam},
  journal={IEEE Trans. Commun.}, 
  title={End-to-End Learning-Based Full-Duplex Amplify-and-Forward Relay Networks}, 
  year={2023},
  volume={71},
  number={1},
  pages={199-213},
  doi={10.1109/TCOMM.2022.3225460}}

@ARTICLE{Chen2023_2,
  author={Chen, Jiechen and Skatchkovsky, Nicolas and Simeone, Osvaldo},
  journal={IEEE Wireless Commun. Lett.}, 
  title={Neuromorphic Integrated Sensing and Communications}, 
  year={2023},
  volume={12},
  number={3},
  pages={476-480},
  doi={10.1109/LWC.2022.3231388}}

@ARTICLE{Eshraghian2023,
  author={Eshraghian, Jason K. and Ward, Max and Neftci, Emre O. and Wang, Xinxin and Lenz, Gregor and Dwivedi, Girish and Bennamoun, Mohammed and Jeong, Doo Seok and Lu, Wei D.},
  journal={Proceedings of the IEEE}, 
  title={Training Spiking Neural Networks Using Lessons From Deep Learning}, 
  year={2023},
  volume={111},
  number={9},
  pages={1016-1054},
  doi={10.1109/JPROC.2023.3308088}}

@ARTICLE{Cammerer2020,
  author={Cammerer, Sebastian and Aoudia, Fayçal Ait and Dörner, Sebastian and Stark, Maximilian and Hoydis, Jakob and ten Brink, Stephan},
  journal={IEEE Trans. Commun.}, 
  title={Trainable Communication Systems: Concepts and Prototype}, 
  year={2020},
  volume={68},
  number={9},
  pages={5489-5503},
  doi={10.1109/TCOMM.2020.3002915}}

@ARTICLE{Gupta2021_2,
  author={Gupta, Ankit and Sellathurai, Mathini},
  journal={IEEE Access}, 
  title={End-to-End Learning-Based Framework for Amplify-and-Forward Relay Networks}, 
  year={2021},
  volume={9},
  number={},
  pages={81660-81677},
  doi={10.1109/ACCESS.2021.3085901}}

@inproceedings{Fang2021,
 author = {Fang, Wei and Yu, Zhaofei and Chen, Yanqi and Huang, Tiejun and Masquelier, Timoth\'{e}e and Tian, Yonghong},
 booktitle = {Advances in Neural Information Processing Systems},
 pages = {21056--21069},
 publisher = {Curran Associates, Inc.},
 title = {Deep Residual Learning in Spiking Neural Networks},
 volume = {34},
 year = {2021}}

@article{3GPP,
  author = {{3GPP}},
  title={Study on channel model for frequencies from 0.5 to 100 {GHz} ({3GPP TR} 38.901 version 16.0.0 Release 16)},
  publisher = {{ETSI}, Sophia Antipolis Cedex, France},
  year={Oct. 2019}}

@article{3GPP38843,
  author = {{3GPP}},
  title={Study on artificial intelligence {(AI)}/machine
learning {(ML)} for {NR} air interface ({3GPP} {TR} 38.843 Release-18 v 2.0.1 )},
  publisher = {ETSI, Sophia Antipolis Cedex, France},
  year={Jan. 2024}}

@article{sionna,
 title = {Sionna: An Open-Source Library for Next-Generation Physical Layer Research},
 author = {Jakob Hoydis and Sebastian Cammerer and Fayçal Ait Aoudia and Avinash Vem and Nikolaus Binder and Guillermo Marcus and Alexander Keller},
 year = {2022},
 month = {Mar.},
 journal = {arXiv preprint},
 online = {https://arxiv.org/abs/2203.11854}
}

@misc{loshchilov2019decoupledweightdecayregularization,
      title={Decoupled Weight Decay Regularization}, 
      author={Ilya Loshchilov and Frank Hutter},
      year={2019},
      eprint={1711.05101},
      archivePrefix={arXiv},
      primaryClass={cs.LG},
      url={https://arxiv.org/abs/1711.05101}, 
}

@inproceedings{nagel2022overcoming,
  title={Overcoming oscillations in quantization-aware training},
  author={Nagel, Markus and Fournarakis, Marios and Bondarenko, Yelysei and Blankevoort, Tijmen},
  booktitle={International Conference on Machine Learning},
  pages={16318--16330},
  year={2022},
  organization={PMLR}
}

@inproceedings{kim2023exploring,
  title={Exploring temporal information dynamics in spiking neural networks},
  author={Kim, Youngeun and Li, Yuhang and Park, Hyoungseob and Venkatesha, Yeshwanth and Hambitzer, Anna and Panda, Priyadarshini},
  booktitle={Proceedings of the AAAI Conference on Artificial Intelligence},
  volume={37},
  number={7},
  pages={8308--8316},
  year={2023}
}

@article{Afshar2014,
  author       = {Afshar, Saeed and George, Libin and Thakur, Chetan Singh and Tapson, Jonathan and van Schaik, André and de Chazal, Philip and Hamilton, Tara Julia},
  title        = {Turn Down that Noise: Synaptic Encoding of Afferent {SNR} in a Single Spiking Neuron},
  journal      = {IEEE Trans. Biomed. Circuits Syst.}, 
  volume       = {9},
  year         = {2015},
  number        ={2},
  pages={188-196},
}

@article{malcolm2023,
  title={A comprehensive review of spiking neural networks: Interpretation, optimization, efficiency, and best practices},
  author={Malcolm, Kai and Casco-Rodriguez, Josue},
  journal={arXiv preprint arXiv:2303.10780},
  year={2023}
}

@INPROCEEDINGS{Horowitz2014,
  author={Horowitz, Mark},
  booktitle={2014 IEEE International Solid-State Circuits Conference Digest of Technical Papers (ISSCC)}, 
  title={1.1 Computing's energy problem (and what we can do about it)}, 
  year={2014},
  volume={},
  number={},
  pages={10-14},
  doi={10.1109/ISSCC.2014.6757323}}

@ARTICLE{Xie2022,
  author={Xie, Kan and Zhang, Zhe and Li, Bo and Kang, Jiawen and Niyato, Dusit and Xie, Shengli and Wu, Yi},
  journal={IEEE Trans. Veh. Tech.}, 
  title={Efficient Federated Learning With Spike Neural Networks for Traffic Sign Recognition}, 
  year={2022},
  volume={71},
  pages={9980-9992},
  doi={10.1109/TVT.2022.3178808}}

@article{Friedemann2021,
    author = {Zenke, Friedemann and Vogels, Tim P.},
    title = {The Remarkable Robustness of Surrogate Gradient Learning for Instilling Complex Function in Spiking Neural Networks},
    journal = {Neural Computation},
    volume = {33},
    number = {4},
    pages = {899-925},
    year = {2021},
    month = {03},
    issn = {0899-7667},
    doi = {10.1162/neco_a_01367},
}

@article{Datta2021,
  title={{HYPER-SNN}: Towards energy-efficient quantized deep spiking neural networks for hyperspectral image classification},
  author={Gourav Datta and Souvik Kundu and Akhilesh R. Jaiswal and Peter A. Beerel},
  journal={arXiv preprint arXiv:2107.11979},
  year={2021}
}

@inproceedings{Dakic2024_2,
  title={{Spiking-UNet}: Spiking Neural Networks for Spectrum Occupancy Monitoring},
  author={Dakic, Kosta and Homssi, Bassel Al and Al-Hourani, Akram},
  booktitle={Proceedings of the  IEEE Wireless Communications and Networking Conference (WCNC)},
  volume={Dubai, United Arab Emirates},
  number={},
  pages={1--6},
  year={2024}
}

@INPROCEEDINGS{Tobias2017,
  author={Gruber, Tobias and Cammerer, Sebastian and Hoydis, Jakob and Brink, Stephan ten},
  booktitle={2017 51st Annual Conference on Information Sciences and Systems (CISS)}, 
  title={On deep learning-based channel decoding}, 
  year={2017},
  volume={},
  number={},
  pages={1-6},
  doi={10.1109/CISS.2017.7926071}}

@ARTICLE{Jialong2023,
  author={Xu, Jialong and Tung, Tze-Yang and Ai, Bo and Chen, Wei and Sun, Yuxuan and Gündüz, Deniz},
  journal={IEEE Communications Magazine}, 
  title={Deep Joint Source-Channel Coding for Semantic Communications}, 
  year={2023},
  volume={61},
  number={11},
  pages={42-48},
  doi={10.1109/MCOM.004.2200819}}

@article{liu2021sstdp,
  title     = {{SSTDP}: Supervised Spike Timing Dependent Plasticity for Efficient Spiking Neural Network Training},
  author    = {Liu, Fan and Zhao, Wenxuan and Chen, Yuxuan and Wang, Zhe and Yang, Tian and Jiang, Li},
  journal   = {Frontiers in Neuroscience},
  volume    = {15},
  pages     = {756876},
  year      = {2021},
  doi       = {10.3389/fnins.2021.756876},
  publisher = {Frontiers}
}

@ARTICLE{Xiao2025,
  author={Han, Xiao and Wenqiang, Tian and Shi, Jin and Wendong, Liu and Jia, Shen and Zhihua, Shi and Zhi, Zhang},
  journal={China Communications}, 
  title={Interference cancellation based neural receiver for superimposed pilot in multi-layer transmission}, 
  year={2025},
  volume={22},
  number={1},
  pages={75-88},
  doi={10.23919/JCC.ja.2024-0220}}

@INPROCEEDINGS{Li2025,
  author={Li, Xinjie and Zhou, Xingyu and Zhang, Jing and Wen, Chao-Kai and Jin, Shi},
  booktitle={2025 IEEE Wireless Communications and Networking Conference (WCNC)}, 
  title={{AI}-Driven Iterative Receiver for Superimposed Pilot Schemes in {MIMO-OFDM} Systems}, 
  year={2025},
  volume={},
  number={},
  pages={1-6},
  doi={10.1109/WCNC61545.2025.10978568}}

\end{document}